\pdfoutput=1
\documentclass{article}
\usepackage{epstopdf}

\usepackage{microtype}
\usepackage{graphicx}
\usepackage{booktabs} 
\usepackage{multirow}
\usepackage[table,xcdraw]{xcolor} 
\usepackage{subcaption}
\usepackage{algorithm}
\usepackage{enumerate}
\usepackage{algpseudocode}
\definecolor{nick_orange}{RGB}{255, 127, 80}
\definecolor{rewardcolor}{RGB}{204, 95, 85} 
\definecolor{reasoncolor}{RGB}{147, 88, 224} 
\definecolor{responsecolor}{RGB}{43, 139, 131} 
\definecolor{promptcolor}{RGB}{244, 153, 94}  
\definecolor{mydred}{RGB}{253, 243, 254}

\definecolor{runpei_orange}{HTML}{F35F27}

\usepackage{enumitem}

\usepackage{hyperref}

\newcommand{\eg}{\textit{e.g.}}
\usepackage[accepted]{icml2025}

\usepackage{amsmath}
\usepackage{amssymb}
\usepackage{mathtools}
\usepackage{amsthm}

\usepackage[capitalize,noabbrev]{cleveref}

\theoremstyle{plain}

\theoremstyle{definition}

\theoremstyle{remark}

\usepackage[textsize=tiny]{todonotes}
\usepackage[utf8]{inputenc}

\icmltitlerunning{Perception in Reflection}

\begin{document}

\twocolumn[
\icmltitle{Perception in Reflection}

\icmlsetsymbol{equal}{*}
\begin{center}
    \begin{minipage}{0.9\textwidth} 
        \centering
        {\bf Yana Wei}$^{*1}$ \hspace{0.4em}{\bf Liang Zhao}$^{*2}$ \hspace{0.4em}{\bf Kangheng Lin}$^3$\hspace{0.4em} {\bf En Yu}$^4$ \hspace{0.4em}{\bf Yuang Peng}$^5$\hspace{0.4em}
        {\bf Runpei Dong}$^6$ \hspace{0.4em}{\bf Jianjian Sun}$^2$ \\{\bf Haoran Wei}$^2$\hspace{0.4em} {\bf Zheng Ge}$^2$\hspace{0.4em} {\bf Xiangyu Zhang}$^2$\hspace{0.4em} {\bf Vishal M. Patel}$^1$ \\
        
        $^1$Johns Hopkins University \quad
        $^2$StepFun \quad 
        $^3$BUPT \quad 
        $^4$HUST \quad 
        $^5$Tsinghua University \quad
        $^6$UIUC \\ 

    \end{minipage}
\end{center}




\vskip 0.4in
]




\begin{abstract}

We present a \textit{perception in reflection} paradigm designed to transcend the limitations of current large vision-language models (LVLMs), which are expected yet often fail to achieve perfect perception initially. Specifically, we propose Reflective Perception (\textbf{RePer}), a dual-model reflection mechanism that systematically alternates between policy and critic models, enables iterative refinement of visual perception. This framework is powered by Reflective Perceptual Learning (\textbf{RPL}), which reinforces intrinsic reflective capabilities through a methodically constructed visual reflection dataset and \textit{reflective unlikelihood training}. Comprehensive experimental evaluation demonstrates RePer's quantifiable improvements in image understanding, captioning precision, and hallucination reduction. Notably, RePer achieves strong alignment between model attention patterns and human visual focus, while RPL optimizes fine-grained and free-form preference alignment. These advancements establish perception in reflection as a robust paradigm for future multimodal agents, particularly in tasks requiring complex reasoning and multi-step manipulation.
\end{abstract}    
\section{Introduction}
\label{sec:intro}

\begin{figure}[t]
\centering
\includegraphics[width=1.0\linewidth]{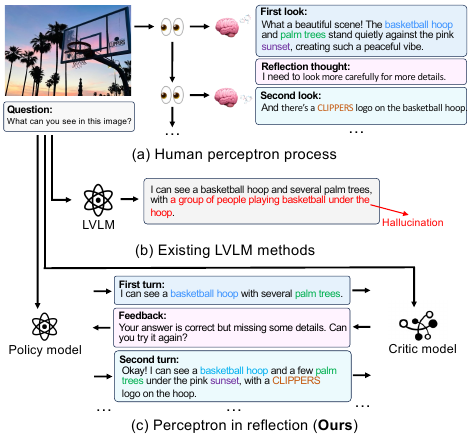}
\caption{Existing LVLMs are expected to deliver accurate perceptions initially, but humans often reflect and refine answers gradually. We introduce \textbf{perception in reflection}, employing policy and critic model interactions to fully harness perceptual capabilities.}
\label{fig:1}
\end{figure}

In advancing large vision-language models (LVLMs)~\citep{gpt4o,llava,qwenvl}, considerable attention has often been focused on enhancing the models’ visual \textbf{perception} capabilities for image understanding. 
This emphasis stems from a fundamental assumption that \textit{well-trained models can achieve sufficiently accurate initial perception}. Such perceptual accuracy enables the model to process visual inputs and generate appropriate responses in a \textit{single pass}~\cite{llava, llava1p5, qwen2vl}.
However, the frequent occurrence of hallucinations and misperceptions hinders their wider applicability in real-world scenarios. 
As shown in~\cref{fig:1}, even for simple scenes, models may generate hallucinatory descriptions (e.g., as seen in (b)) or fail to capture essential details (e.g., as observed in the initial response in (c)).
This raises an important consideration: \textit{Are current perception paradigms inherently limited, or might there be a more reasonable paradigm?}

Some methods~\citep{shikra,chartthinker,merlin} attempt to mitigate this through a sort of visual chain-of-thought (CoT)~\citep{cot} reasoning. They establish a paradigm that first executes fine-grained perceptual tasks (such as grounding object locations~\citep{shikra,visualcot}, structures~\citep{chartthinker} or identities~\citep{merlin}) before engaging in broader perception. 
However, these approaches face a key limitation: the reliance on specialized tasks and data formats that are difficult to generalize across all vision-language tasks, \eg, box CoT can not be used in math geometry problems, making it challenging to achieve consistent visual perception across diverse scenarios. Furthermore, CoT does not change the original single-pass manner. When perceptual errors occur, it is unable to adjust and rectify them.


Shifting the view to the real world, we can observe that humans, as shown in~\cref{fig:1}, typically do not perceive in a single step, rather, they establish cognition through gradual observation. This iterative process enables humans to continually enrich, refine, and enhance their perceptual outcomes. Drawing inspiration from this, we think that \textit{a reasonable perception paradigm for LVLMs should be iterative rather than a single-pass.} In other words, the ability to reflect and improve over multiple rounds is not just a desirable feature; it’s a \textit{fundamental requirement} for LVLMs to achieve robust and generalizable perception.


In this paper, we propose a novel perceptual mechanism, termed \textbf{Re}flective \textbf{Per}ception (\textbf{RePer}). Its purpose is to enable LVLMs to, like humans, use a perception-feedback loop to gradually establish precise visual cognition. To achieve this, we make RePer a dual-model architecture, \textit{i.e.}, \textit{policy} model and \textit{critic} model, to enable LVLMs to conduct \textit{percption} and \textit{reflection} separately in terms of multi-turn dialogues between policy and critic model. In this way, LVLMs distill lessons from past experiences, gradually direct attention toward critical visual primitives, and thereby produce more accurate and refined responses.

Although LVLMs inherently possess reflective perception capabilities, this ability is instable and has not been effectively activated~\citep{score}. To this end, we further propose a \textbf{R}eflective \textbf{P}erceptual \textbf{L}earning (\textbf{RPL}) approach.
Through strategic temperature sampling and a hybrid evaluation system combining model and rule-based rewarding, we construct an online, multi-turn \textit{visual reflection dataset}. This dataset exhibits progressive improvements in both perception accuracy and response quality across dialogue turns. Building upon this, we propose \textit{reflective unlikelihood training}, an imitation learning approach~\citep{imitation,swamy2023inverse} that calibrates the model’s preferences across responses of varying quality, thereby mitigating \textit{behavioral collapse}~\citep{score} where models tend to generate suboptimal responses in early turns.

Extensive experiments demonstrate that RePer achieves superior performance across various benchmarks including image understanding, hallucination detection and detailed image caption, \eg, $54\%$ CAPTURE on DetailCaps~\citep{dong2024benchmarking} and $51\%$ accuracy on HallusionBench~\citep{guan2024hallusionbench}. Using GPT-4o~\citep{gpt4o} and DALLE-3~\citep{dalle3}, we validate its enhanced perception capabilities from both discriminative and generative perspectives. Comprehensive ablation studies on data construction, training strategies, reflection rounds, and critic designs verify RePer’s generalizability, establishing it as a fundamental paradigm for advancing multimodal perception.

In order to thoroughly unveil the underlying mechanisms behind \textit{perception in reflection}, we further conducted a series of analytical experiments. Our comprehensive experimental analysis reveals two key findings:

\begin{itemize}[leftmargin=2.5mm]
\setlength{\itemsep}{2pt}

    \item RePer can effectively \textit{migrate image attention towards human-aligned regions} through iterative refinement. This implies that the perceptual pattern utilized by RePer aligns more closely with that of humans.
    
    \item RPL can be regarded as a \textit{free-form preference optimization} framework that unifies various preference learning paradigms, \eg, DPO~\cite{dpo}, and LiPO~\citep{lipo}, while enabling fine-grained supervision through explicit feedback signals.  
    
\end{itemize}
\vspace{-3mm}
These two key findings underscore the crucial value of \textit{perception in reflection} in enhancing multimodal understanding and reasoning capabilities. We believe it will become an essential capability for multimodal agents in the future, particularly in complex visual reasoning~\citep{agent1,visualreason} and multi-step manipulation~\citep{vla1,vla2} tasks.

\section{Perception in Reflection}

In this section, we first define our problem and formalize the objective from a reinforcement learning perspective (\cref{sec:problem_formulation}). We then elaborate on how models learn to perceive through reflection, encompassing both data construction and training strategies (\cref{sec:rpl}). Finally, we present the inference algorithm for reflective perception during deployment (\cref{sec:rp-inference}).

\subsection{Problem Definition and Formulation}
\label{sec:problem_formulation}

\textbf{Perception in LVLMs.} Perception, as a concept in the field of computer vision~\cite{resnet, fasterrcnn, maskrcnn}, refers to the process of interpreting and understanding sensory, \textit{ie.,} vision, information from the environment. In the context of LVLM, we typically define perception as the process by which the model recognizes and understands the image or video. The perception capability of the model will directly determine the accuracy of its understanding and reasoning towards real world.

\textbf{Perception in Reflection.} Our goal is to mimic human perception, establishing a perceive-feedback loop through LVLM’s iterative attempts to enhance image comprehension and response accuracy. In pursuit of this, we model our challenge through the lens of reinforcement learning (RL), inspired by SCoRe~\citep{score} and RISE~\citep{recursiveintro}. To be specific, given a dataset $\mathcal{D} = \{(I_i, x_i, y_i^*)\}_{i=1}^{N}$ of images $I_i$, questions $x_i$, and oracle responses $y_i^*$, we aim to train an LVLM policy $\pi_\theta(\cdot \mid [I, x, \hat{y}_{1:t}, f_{1:t}])$. This model, given an image $I$ and question $x$, along with $t$ previous attempts $\hat{y}_{1:t}$ and feedback prompts $f_{1:t}$, is designed to perceive the image as accurate as possible and deliver the most correct possible answer $y$. Formally, given a verifier $r(y, y^*)$ to assess the correctness of model response $y$ compared to oracle answer $y^*$, we aim to derive a policy that utilizes the aforementioned information to produce the outputs with the highest correctness reward over $T$ rounds:
\begin{equation}
    \max_{\pi_\theta} \sum_{t=1}^{T} \mathbb{E}_{I, x, y^* \sim \mathcal{D}, \hat{y}_{t} \sim \pi_\theta(\cdot \mid [I, x, \hat{y}_{1:t-1}, f_{1:t-1}])} r(\hat{y}_t, y^*).
\end{equation}\label{eq:obj}
\cref{eq:obj} resembles a multi-round Markov Decision Process (MDP)~\citep{recursiveintro} or can be viewed as an RL or supervised finetune (SFT) objective. It is noteworthy that every historical attempt is synchronously optimized to maximize the ultimate reward.

\subsection{Reflective Perceptual Learning}
\label{sec:rpl}

Despite existing LVLMs often possessing intrinsic self-reflection capabilities~\citep{selfcorrect}, these abilities have been shown to be remarkably fragile~\citep{score}. In other words, they struggle to adaptively refine their responses based on given feedback (as shown in \cref{fig:case-llava-multiturn}). To address this limitation, we propose Reflective Perception Learning (RPL), a methodology that trains models to continuously enhance their previous responses through imitation learning~\citep{imitation,swamy2023inverse}. We first elaborate on the data collection and training objective.

\begin{figure}[t]
\centering
\includegraphics[width=1.0\linewidth]{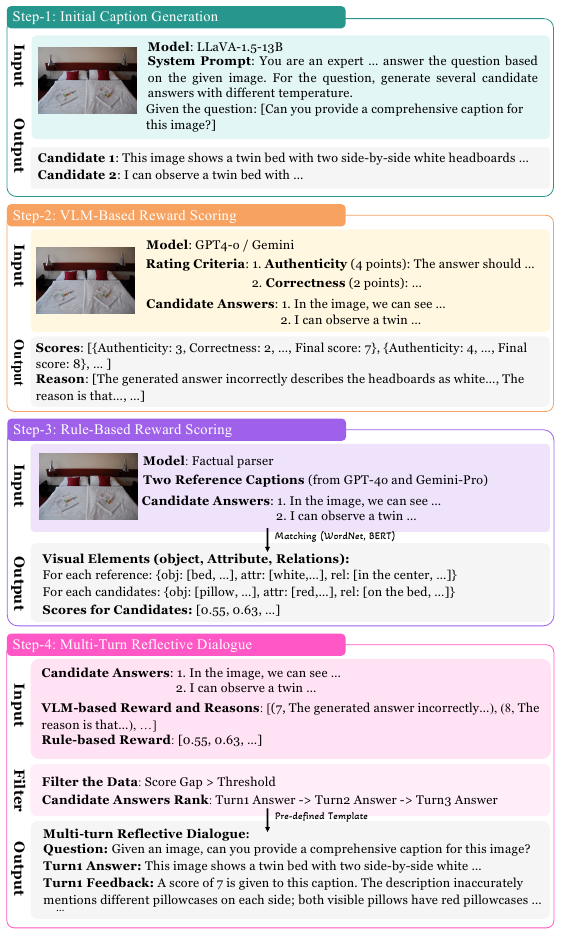}
\caption{\textbf{Data construction pipeline of visual reflection dataset}.}
\label{fig:data_construction}
\end{figure}

\noindent\textbf{Data Construction.}
Naturally, we structure a multi-turn dialogue encompassing the sequence of posing questions, providing responses, receiving erroneous feedback, and subsequently re-responding and re-evaluating. This iterative process is designed to cultivate and demonstrate reflective perception capabilities within the trained models. 

Practically, we expect the model to, 
\textbf{(1)} generate \textit{diverse} responses based on all past answers and feedback, thereby enabling the \textit{exploration} during reflection towards a perception with fewer errors; 
\textbf{(2)} gradually produce more \textit{accurate} answers in multi-turn dialogues, ensuring the \textit{convergence} of the reflective chain. 
To meet these requirements, we construct a visual reflection dataset for model imitation. 
\cref{fig:data_construction} gives an overview, with detailed steps as follows:

\textbf{\textit{Step-1}: Initial Candidate generation.} We employ \textit{temperature sampling} to generate diverse candidate answers per image-question pair. This approach ensures sufficient variation in response style, detail level, and accuracy while maintaining semantic relevance.

\textbf{\textit{Step-2} VLM-Based Reward Scoring.} For the generated multiple candidate responses, we employ a robust Visual-Language Model (VLM) to conduct a comprehensive and multifaceted evaluation, yielding fine-grained scores.

\textbf{\textit{Step-3} Rule-Based Reward Scoring.} Then we design a pipeline to extract key elements, \eg, objects, attributes, and relations, from both images and responses, and establish matching rules to compute alignment scores.

\textbf{\textit{Step-4} Reflective Dialogue construction.} After obtaining the candidate answers and their corresponding reward scores, we select samples meeting two criteria: (a) a minimum score gap between the highest and lowest responses, and (b) at least one response scoring above the specified points. Then the filtered responses are structured into $N$ rounds of reflective dialogue, progressing from lowest to highest scores. To this end, we curate a dataset $\tilde{\mathcal{D}} = \{\{(I^i_t, x^i_t, \tilde{y}^i_t,f^i_t,r^i_t )\}_{t=1}^{T}\}_{i=1}^{N}$, where $\tilde{y}^i_t$ is sampled from model outputs, $f^i_t$ represents specific feedback, and $r^i_t$ denotes the corresponding reward score.

\begin{algorithm}
\caption{Reflective Perception (RePer)}
\label{alg}
\begin{algorithmic}[1]
\State Initialize Policy, Critic model: $\pi_\theta, r_\theta$
\State Generate initial perception response $y_0$ using $\pi_\theta$ given image $I$ and language instruction $x$
\State Generate initial evaluation $r_0, f_0$ using $r_\theta$ given ($I$, $x$, $y_0$)
\State Set $t \gets 0$
\While{$t < \text{max trials}$}
    \State Generate perception response $y_t$ using $\pi_\theta$ given ($I, x$, $y_0, r_0, f_0, ..., y_{t-1}, r_{t-1}, f_{t-1}$)
    \State Generate evaluation $r_t, f_t$ using $r_\theta$ given ($I, x$, $y_0, r_0, f_0, ..., y_{t-1}, r_{t-1}, f_{t-1}, y_t$)
    \State Increment $t$
\EndWhile
\State \textbf{return}
\end{algorithmic}
\end{algorithm}

Two points merit attention. First, it is crucial to reward answer of each round using a hybrid scoring mechanism. This approach aims to align the model with both \textit{rule-based} and \textit{model-based} reward systems~\citep{mu2024rule}, thereby maximizing its ability to generalize to complex real-world scenarios.
Second, we aim to devise responses based on the self-generated outputs of the model, thereby facilitating an \textit{online} optimization process. This is intended to minimize the risk of the model \textit{overfitting} to non-reflective capabilities~\citep{score,recursiveintro,online}.

\noindent\textbf{Reflective Unlikelihood Training.}
Based on the constructed data, we apply imitation learning~\citep{imitation,swamy2023inverse} to simulate reflective perception. This learning process necessitates the disregard of textual patterns, focusing instead on the cultivation of capabilities.

More critically, we seek to prevent the model from overfitting to multi-turn responses and avoid the \textit{behavioral collapse}~\citep{score} where the model consistently generates \textit{suboptimal} initial replies. In previous efforts, both RISE~\citep{recursiveintro} and SCoRe~\citep{score} primarily utilized SFT for imitation learning. However, RISE employed the exponent of \textit{centered rewards} to mitigate this issue, while SCoRe utilized \textit{reward shaping} to counteract. In this paper, we propose a method that simultaneously balances likelihood and unlikelihood~\citep{unlikelihood}, formalized as follows:
\begin{equation}\small\label{eq:policy}
\max_{\theta} \mathbb{E}_{\circ^i \sim \tilde{D}} { \sum_{t=1}^{T} \sigma_t^i \log \pi_{\theta}(\tilde{y}_t^i | \circ_t^i) + \alpha(1-\sigma_t^i) \log(1- \pi_{\theta}(\tilde{y}_t^i | \circ_t^i))},
\end{equation}
where $\circ$ denotes a single sampling instance from our constructed dataset $\tilde{\mathcal{D}}$, and $\sigma_t^i = F(r_t^i)$ represents the normalization of reward $r_t^i$. $\alpha$ is a constant term that adjusts the unlikelihood loss scale.

Essentially, we employ rewards to balance likelihood and unlikelihood. In the initial rounds where the reward is lower (smaller loss weight), there is a predisposition towards unlikelihood, promoting the penalization of the response. Conversely, in subsequent rounds where the reward is higher (larger loss weight), there is a tendency towards likelihood, encouraging rewarding of the response. We will elaborate this on~\cref{sec:fpo}.

\subsection{Reflective Perception}
\label{sec:rp-inference}

\begin{figure}[t]
\centering
\includegraphics[width=1.0\linewidth]{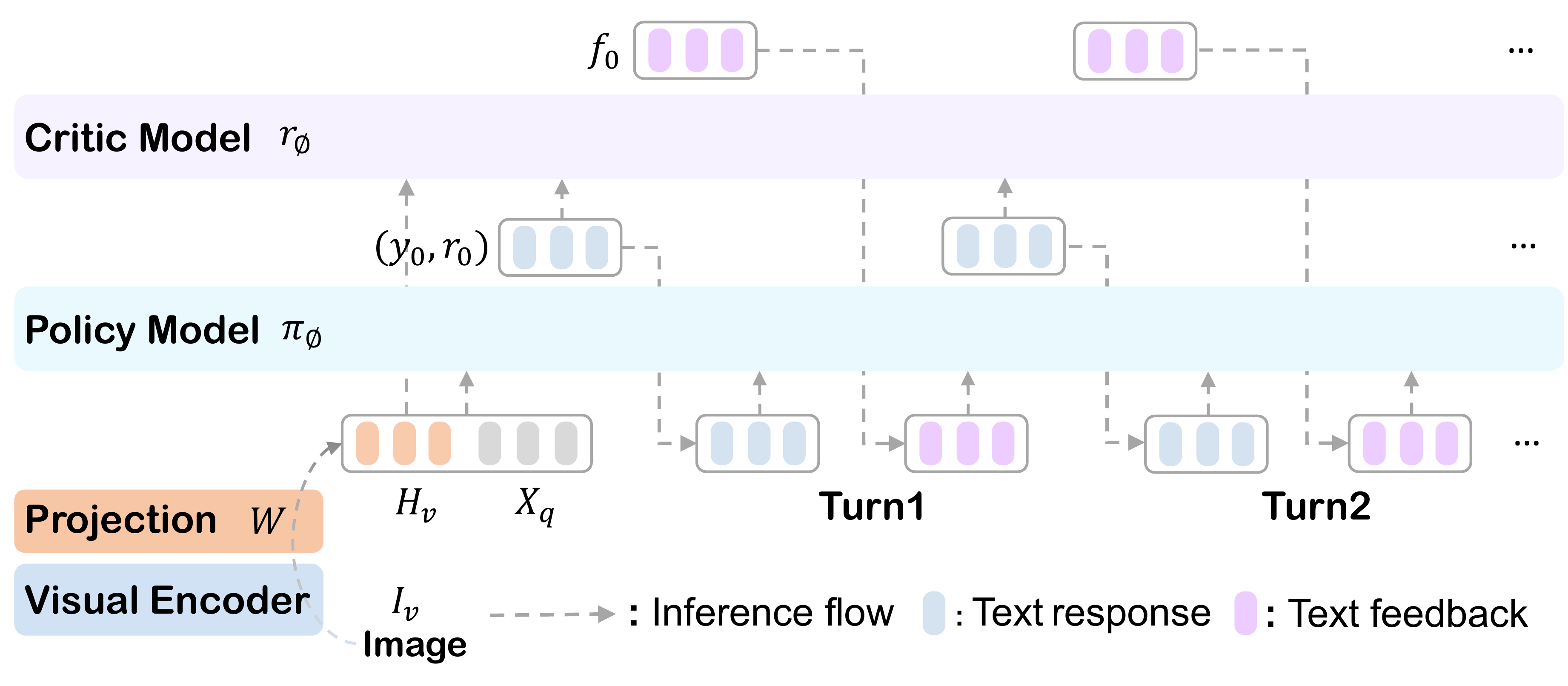}
\caption{\textbf{Inference pipeline of reflective perception}.}
\label{fig:3}
\vspace{-6mm}
\end{figure}

As shown in~\cref{alg} and~\cref{fig:3}, we define reflective perception during the inference process as a \textit{collaborative interaction} between the well-trained policy and critic agents.

Initially, the policy model observes and provides its perceptual results, which are then evaluated by the critic model. The critic model assesses the policy’s perception, providing both a score to gauge the quality of the perception and a rationale for the score given. Following this, the policy model reflects on its perceptual errors, informed by the critic’s evaluations on self-generated responses from previous rounds, and produce a new perceptual response. The critic model then delivers updated feedback based on all previous perceptions and critiques.
This iterative process continues until a predetermined limit is reached.

\begin{figure*}[ht]
\centering
\includegraphics[width=0.8\linewidth]{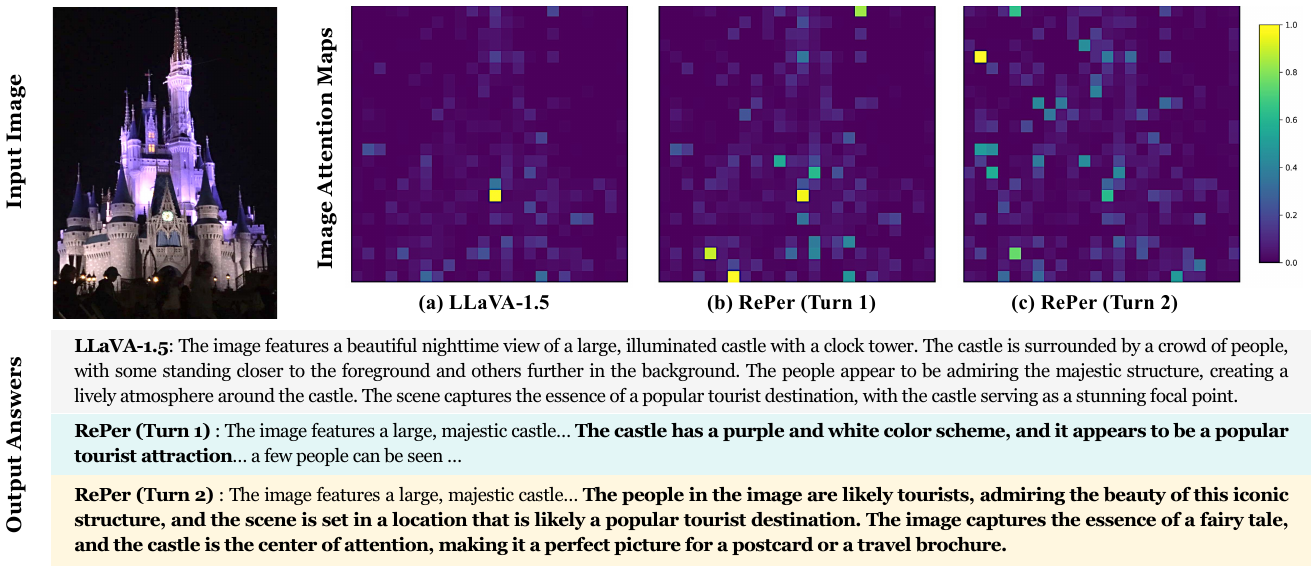}
\caption{\textbf{Comparison of image attention maps between LLaVA-1.5 and RePer}, highlighting RePer’s broader activation of image tokens and its ability to generate more detailed and accurate answers. While LLaVA-1.5 over-focuses on “people”, RePer correctly attends to the main subject, “castle,” progressively activating more relevant tokens for improved perception.}
\label{fig:attn_compare}
\end{figure*}

Through an iterative reflective perception mechanism, the model can achieve human-like visual perception patterns, as empirically validated by the experimental results demonstrated in Figure~\ref{fig:attn_compare}.

\section{Discussion}



\subsection{RePer Progressively Aligns Human Attention.}\label{sec:attn}

The reflective capabilities of language models have been widely studied~\citep{selfcorrect,score,recursiveintro}. It is suggested that these models inherently possess a reflective ability that, although unstable, can improve the certainty of outputs and activate deeper levels of knowledge~\citep{selfcorrect}. Yet, the effectiveness of this capacity in LVLMs remains an unexplored question.

To investigate, we begin by visualizing the detailed image attention of the model for each round of the model’s responses.
As shown in~\cref{fig:attn_compare}, as the rounds progress, the model \textit{accurately shifts its attention emphasis} towards the correct image tokens, rather than over-focusing on a few insignificant ones. Hence correspondingly, the hallucinations and erroneous perceptions initially displayed by the model gradually decrease. 
Essentially, this represents a \textbf{\textit{progressive alignment towards ground-truth human attention}}.

\cref{fig:attn-increase} provides further quantitative support, showing a log-linear increase in average image token activations over five rounds of reflection. Specifically, we calculated the average image token activations across 1,000 cases to measure how the model’s attention to images varies during reflective perception. This is important because responses with fewer hallucinations are associated with higher average activations of image tokens~\citep{opera}.
Our findings suggest that visual reflection \textit{gradually unlocks the model’s inherent visual capabilities}, focusing attention on salient image context and progressively mitigating hallucination.

\subsection{RPL is a Free-Form Preference Optimization.}\label{sec:fpo}

Revisiting the data construction in RPL, we essentially transform \textit{listwise preference data} with precise feedback and scores into multi-turn dialogues grading from poor to good quality. This prompts the inquiry: is RPL fundamentally a preference optimization process?

Revisiting~\cref{eq:policy}, for a given sample $\circ$ and its $T$ dialogue iterations, the objective is articulated as follows:
\begin{equation}\small
\begin{aligned}
L^i = &\underbrace{\sigma_1 \log \pi_{\theta}(\tilde{y}_1 | \circ_1)}_{\text{less likelihood}} + \underbrace{\alpha(1-\sigma_1) \log(1- \pi_{\theta}(\tilde{y}_1 | \circ_1))}_{\text{more unlikelihood}}+...+ \\
&\underbrace{\sigma_T \log \pi_{\theta}(\tilde{y}_T | \circ_T)}_{\text{more likelihood}} + \underbrace{\alpha(1-\sigma_T) \log(1- \pi_{\theta}(\tilde{y}_T | \circ_T))}_{\text{less unlikelihood}}.
\end{aligned}\label{eq:explain}
\end{equation}

As aforementioned, to develop reflective perception capabilities, we create multi-turn data that progresses from poor to good responses, with rewards increasing linearly from rounds $1$ to $T$. As a result, in the initial rounds, the model mainly penalizes poor samples (\textit{more unlikelihood}), while in later rounds, it gradually shifts to rewarding good samples (\textit{more likelihood}). This helps the model avoid overfitting to poor initial samples and, importantly, allows it to progressively learn to distinguish between good and bad samples.

From another perspective, we can view RPL as a form of \textit{reward modeling}. Unlike popular LLM-based reward modeling methods such as DPO~\citep{dpo} and LiPO~\citep{lipo}, RPL does not propagate gradients to the remaining negative samples. Yet, back-propagation over multi-round dialogues is actually not isolated. With each response \textit{contextualizing} all previous responses, as denoted by $\circ_t=[I, x, \boldsymbol{\hat{y}_{1:t-1}}, \boldsymbol{f_{1:t-1}}]$, each sample implicitly establishes a \textbf{\textit{partial increasing preference order}}. 

Moreover, it is worth noting that RPL holds a significant advantage over previous reward modeling approaches: flexibility in handling diverse preference samples—\textit{pairwise or listwise, scalar or fine-grained feedback}-based rewards—while maintaining stable training. Additionally, the use of detailed feedback aids error highlighting, facilitating object-level or even token-level preference that direct optimization more precisely. Our analyses in~\cref{exp_analysis} further confirms this.

\section{Experiments}
\label{exp}


\begin{table*}[htbp]
\centering
\caption{\textbf{Model Performance Comparison of RePer with Baselines and State-of-the-Art Models.} RePer outperforms across six benchmarks, with the best results highlighted in \textbf{bold}.}
\label{tab-exp:ref-perc}
\resizebox{0.95\textwidth}{!}{ 
\begin{tabular}{l|cc|ccc|ccc|c|cc|ccccccc}
\toprule
\textbf{Model} & \multicolumn{2}{c|}{\textbf{MMHal-Bench}} & \multicolumn{3}{c|}{\textbf{HallusionBench}}   & \multicolumn{3}{c|}{\textbf{Detailcaps-4870}} & \textbf{LLaVABench} & \multicolumn{2}{c|}{\textbf{GAIVE}} & \multicolumn{3}{c}{\textbf{GAPE}}\\
               & Score ↑ & Hal rate ↑ & aAcc ↑ & fAcc ↑ & qAcc ↑ & CAPTURE ↑ & Precision ↑ & Recall ↑ & & Relevancy ↑ & Accuracy ↑ & Authen. ↑ & Correct. ↑ & Total ↑\\
\midrule
MiniGPT-4 7B               & -      & -    & 35.78 & 10.12 & 8.79    & -     & -     & -     & 45.1 & -  & -  & -         & -     & -\\
mPLUG-Owl 7B               & -      & -    & 43.93 & 10.40 & 9.45    & -     & -     & -     & -    &  - &  - & -         & -     & -\\
InstructBLIP 7B            & -      & -    & 45.26 & 10.11 & 9.45    & 51.81 & 65.22 & 45.01 & 59.8 & -  & -  & -         & -     & -\\
LLaVA-SFT+ 7B              & 1.88   & 0.68 & 33.65 & 8.96  & 5.93    & 51.13 & 64.38 & 44.28 & 44.6 & 6.68  & 4.85  & 27.62 & 12.47  & 70.09\\
LLaVA-RLHF 7B              & 1.67   & 0.76 & 31.23 & 14.16 & 7.69    & 52.21 & 63.61 & 45.93 & 44.9 & 4.88  & 4.27  & 27.93 & 12.64 & 70.68\\
VOLCANO 7B                 & 2.06   & 0.62 & 26.50 & 10.69 & 6.37    & 50.88 & 66.23 & 43.35 & 54.0 &  7.12 &  5.35 & 31.63 & 14.52 & 78.78\\
LLaVA-SFT+ 13B             & 1.92   & 0.65 & 46.37 & 22.25 & 18.24   & 51.08 & 64.48 & 44.04 & 55.8 &  6.85 & 5.20  & 30.00 & 13.44 & 74.88\\
LLaVA-RLHF 13B             & 2.09   & 0.69 & 36.20 & 15.32 & 14.73   & 52.05 & 64.56 & 45.35 & 62.6 & 4.66  & 4.33  & 30.06 & 13.59 & 75.36\\
VOLCANO 13B                & 2.15   & 0.64 & 40.69 & 19.36 & 13.40   & 51.21 & 66.47 & 43.65 & 66.0 &  7.55 &  5.59 &31.34	&14.32	&78.17\\
\midrule
LLaVA-1.5 7B               & 2.02 & 0.61 & 35.65 & 17.92 & 11.21   & 51.03 & \textbf{67.27} & 42.19 & 60.2 &  6.50 & 5.28  & 30.19 & 13.58 & 75.16\\
\rowcolor{mydred} 
\textbf{ +RePer}            & 2.51     &  0.53    &   38.70    &  19.65     &  14.29       & 52.89 & 66.81 & 45.69 & 60.7 & 6.91 & 6.04 & 33.16 & 14.94 & 80.88  \\
\midrule
LLaVA-1.5 13B              & 2.35 & 0.58 & 43.85 & 20.81 & 14.95   & 51.23 & 66.26 & 43.77 & 66.95 & 6.65  & 5.49  & 31.27 & 14.12 & 77.37\\

\rowcolor{mydred} 
\textbf{ +RePer}           & \textbf{2.61} & \textbf{0.52} & \textbf{51.00} & \textbf{22.83} & \textbf{20.00} & \textbf{54.73} & 64.74 & \textbf{49.1} & \textbf{67.6} &  \textbf{7.67} & \textbf{6.86}  & \textbf{34.11} & \textbf{15.33} & \textbf{82.54}\\
\bottomrule
\end{tabular}
}
\vspace{-3mm}
\end{table*}

\subsection{Implemental Details}
\label{exp_imp}

\noindent \textbf{Datasets.}
To construct the training dataset as illustrated in~\cref{sec:rpl}, we begin by randomly sampling 10,000 images from the LLaVA-665K~\cite{llava} dataset. For each image, we prompt the model to generate 8 different captions sampled with temperatures ranging from 0.0 to 1.4 in increments of 0.2.
To filter high-quality samples, we retain instances from VLM-based scoring where the highest score exceeds 9 and the score disparity (difference between the highest and lowest scores) is greater than 4. Similarly, for rule-based scoring, we retain cases with a highest score above 0.55 and a score disparity exceeding 0.2.
Using the generated captions, rewards, and templates from~\cref{fig:data_construction}, we create the \textbf{visual reflection dataset}, containing 11,065 samples from 8,101 images. These samples are distributed as follows: 3,649 for one conversation turn, 2,621 for two turns, and 3,795 for three turns.

\noindent \textbf{Models Training and Inference.} 
Our experiments are based on the LLaVA-1.5~\citep{llava1p5} architecture. We directly supervised finetune the instruct model on our generated datasets. All models are trained for one epoch on 8 NVIDIA A100 GPUs with a batch size of 8 and a learning rate of 1e-6. Only the parameters of the LLM module are fine-tuned, while the rest remain frozen.
In reflective unlikelihood training (\cref{eq:policy}), rewards are normalized to [0,1] by dividing with their maximum values ($F$), serving as likelihood weight ($\sigma$). The constant term $\alpha$ is set as 10.0.
During the inference stage mentioned in \cref{sec:rp-inference}, we defaultly use LLaVA-Critic~\cite{llavacritic} as the critic model.

\subsection{Main Results}
\label{exp_pr}




To evaluate the visual perception capabilities of RePer, we conducted assessments across five widely-used benchmarks, covering a range of tasks: image understanding (LLaVABench~\cite{llava}), hallucination detection (HallusionBench~\cite{guan2024hallusionbench}, MMHal-Bench~\cite{sun2023aligninglargemultimodalmodels}, GAIVE~\cite{liu2023aligning}), and detailed image captioning (DetailCaps~\cite{dong2024benchmarking}). As shown in~\cref{tab-exp:ref-perc}, we compared RePer not only with classic state-of-the-art multimodal baselines including MiniGPT-4~\cite{minigpt4}, mPLUG-Owl~\cite{mplug}, InstructBLIP~\cite{instructblip}, LLaVA~\cite{llava}, LLaVA-RLHF~\cite{2023llavarlhf}, LLaVA-1.5~\cite{llava1p5} but also with Volcano~\cite{volcano}, a multimodal model trained with self-feedback guided refinement.


As shown in \cref{tab-exp:ref-perc}, RePer consistently outperforms baseline models across benchmarks and model scales. Its notable improvement on DetailCaps (+3.64\% in 7B and +6.83\% in 13B) highlights its ability to generate more accurate and detailed captions through multi-turn refinement and RPL. The increased recall rate (+8.30\% in 7B and +12.17\% in 13B) for visual elements demonstrates RePer’s enhanced perception of details. This results in consistent improvements on general and hallucination-related benchmarks, reducing hallucinations without sacrificing image understanding.

\subsection{GPT-4o-Assisted Perception Evaluation (GAPE)}
\label{exp:gape}



We introduce GPT-4o-Assisted Perception Evaluation (GAPE) to simulate human-like perception assessment. Designed to complement traditional closed-set image captioning benchmarks~\cite{chen2015microsoft, agrawal2019nocaps}, GAPE evaluates model-generated captions by leveraging human-aligned prompts with GPT-4o~\cite{peng2024dreambench++} without the need for human-annotated groundtruth answers. 
Specifically, given an image and a prompt, GPT-4o evaluates the generated captions across five dimensions: \textit{Authenticity}, \textit{Correctness}, \textit{Detail}, \textit{Coherence}, and \textit{Completeness}. The evaluation prompts align with the “Rating Criteria” outlined in~\cref{fig:data-reward-pipe}. To better highlight differences in caption quality, these dimensions are scored on a larger scale from 0 to 100, offering a human-like and nuanced assessment of caption performance.



As shown in~\cref{tab-exp:ref-perc} and~\cref{tab-exp:gape}, our RePer consistently outperforms other methods, demonstrating its effectiveness in enhancing model’s perceptual capabilities. Notably, we observe the most significant improvement in \textit{Authenticity}, which evaluates the model’s tendency to hallucinate non-existent objects. This substantial gain can be attributed to our unlikelihood training objective, which effectively penalizes misaligned visual descriptions.

\subsection{Evaluation via Text-to-Image Reconstruction}
\label{exp:txt2img}
\begin{table}[t]
\centering
\caption{Image captioning comparison on 13B models using the CLIP-Image-Score metric and its variants with DINO/DINOv2 as Image encoders.}
\label{tab:txt2img}
\resizebox{0.35\textwidth}{!}{
\fontsize{5}{6}\selectfont 
\begin{tabular}{lccc}
\toprule[.4pt]
\textbf{Model} & \textbf{CLIP} & \textbf{DINO} & \textbf{DINOv2} \\
\midrule[.3pt]
LLaVA-1.5 & 67.43 & 40.56 & 41.02 \\
\rowcolor{mydred} +RePer & \textbf{67.85} & \textbf{42.19} & \textbf{42.12} \\
\bottomrule[.4pt]
\end{tabular}
}
\end{table}

\begin{figure}[htbp]
\centering
\includegraphics[width=1.0\linewidth]{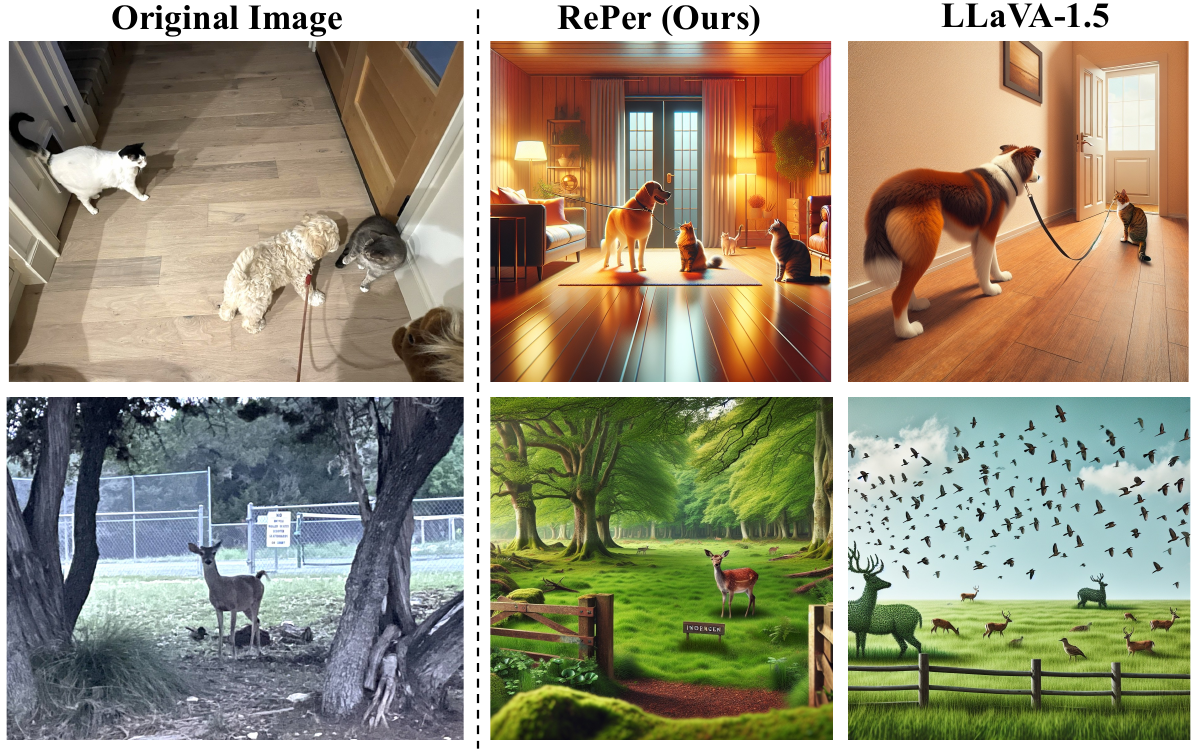}
\caption{We use DALLE-3~\citep{dalle3} as a text-to-image model to reconstruct images using generated captions. Compared to the original image, reconstructed images from LLaVA-1.5~\citep{llava1p5} captions lack key objects or include extraneous ones, indicating incomplete descriptions or hallucinations.}
\label{fig:txt2img}
\vspace{-3mm}
\end{figure}

We assess image captioning performance, a key perceptual application, using the CLIP-Image-Score metric from VisualFactChecker~\citep{ge2024visual}. This metric evaluates caption accuracy and detail by comparing the similarity between an original image and its text-to-image generated version (DALLE-3~\citep{dalle3}), using the caption as a prompt. By comparing the raw and reconstructed images, the metric detects hallucination-related discrepancies, providing a unique perspective on caption quality. To enhance this evaluation, we substitute the CLIP model with DINO~\citep{caron2021emerging} and DINOv2~\citep{darcet2023vitneedreg} for a more thorough assessment. 

As shown in ~\cref{tab:txt2img}, our RePer consistently outperforms the baselines, underscoring the superior quality of its captions. \cref{fig:txt2img} presents visual examples of the reconstruction process. In the second example, LLaVA 1.5 falsely mentions, “There are several birds scattered throughout the scene,” exhibiting hallucination. In contrast, the caption from our RePer produces a reconstructed image closely resembling the original, demonstrating its superior accuracy and ability to avoid hallucinations.

\subsection{Ablation Studies}

\begin{figure*}[ht]
\centering
\begin{subcaptionblock}{0.24\textwidth}
    \includegraphics[width=\linewidth]{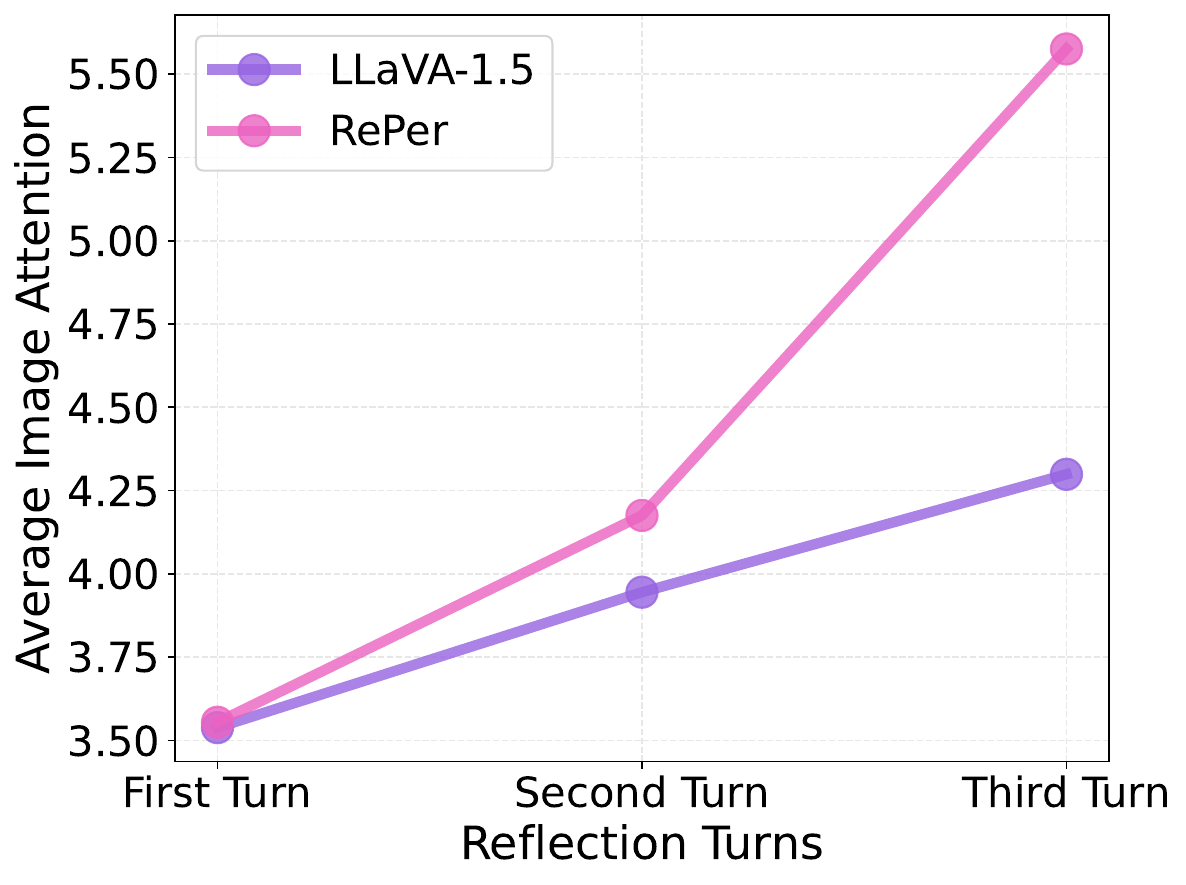}
    \caption{Attentive Img-Token Analysis}
    \label{fig:attn-increase}
\end{subcaptionblock}
\begin{subcaptionblock}{0.24\textwidth}
    \includegraphics[width=\linewidth]{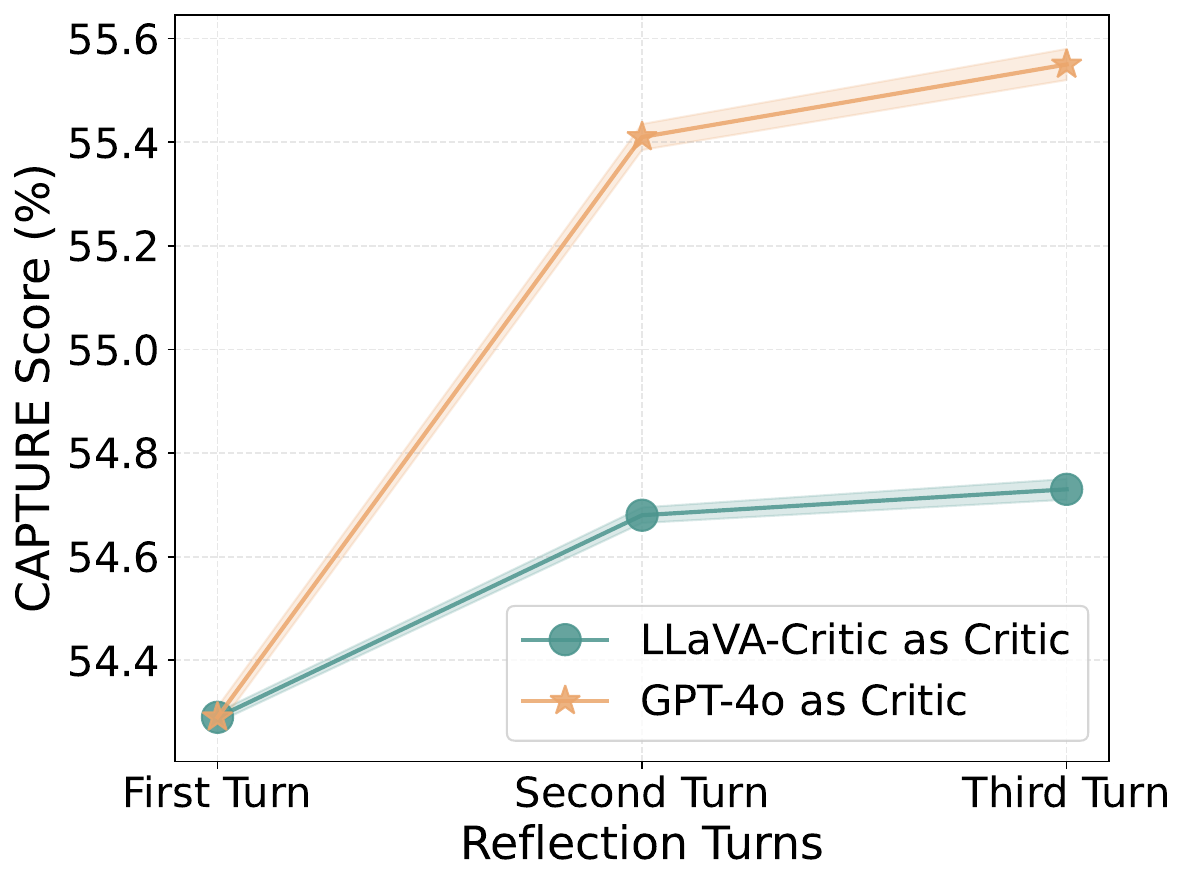}
    \caption{Reflection Turns}
    \label{fig:abl-round}
\end{subcaptionblock}
\begin{subcaptionblock}{0.24\textwidth}
    \includegraphics[width=\linewidth]{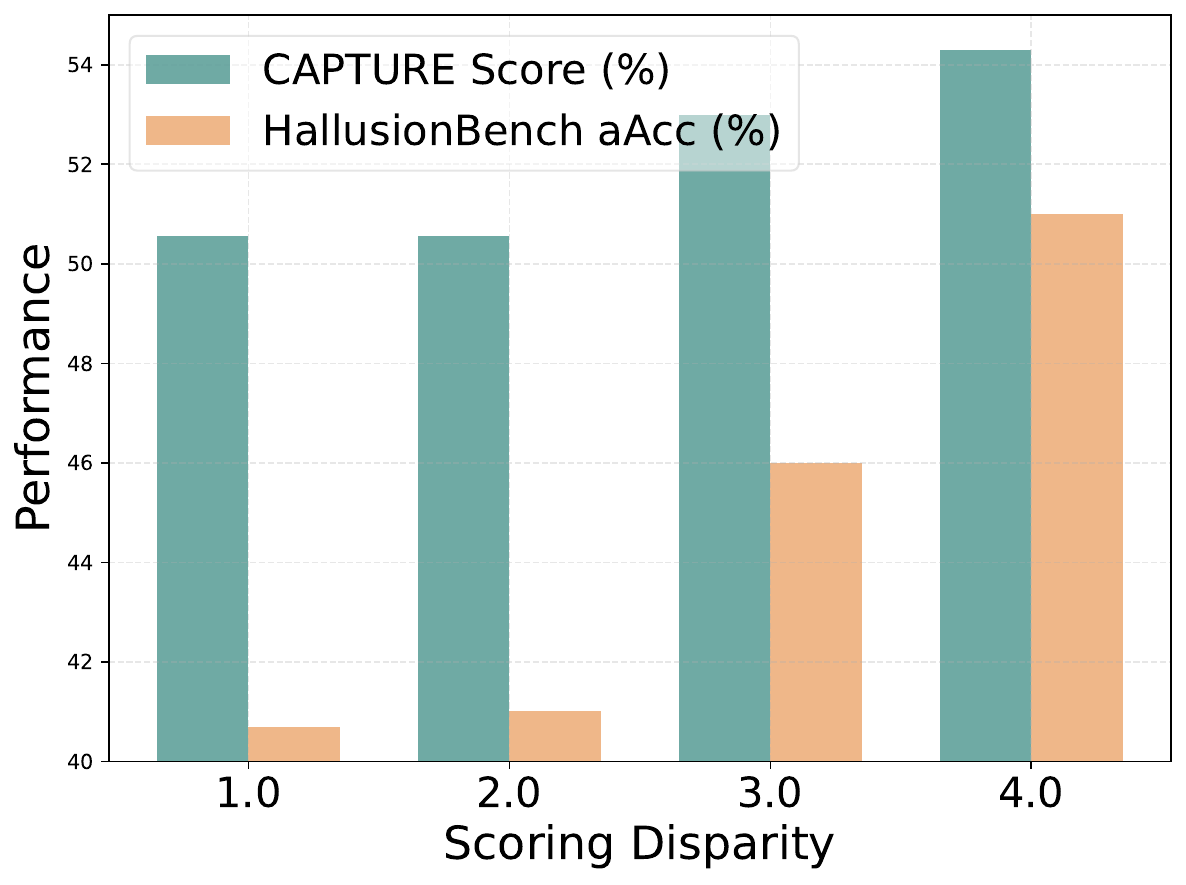}
    \caption{Scoring Disparity}
    \label{fig:abl-disparity}
\end{subcaptionblock}
\begin{subcaptionblock}{0.24\textwidth}
    \includegraphics[width=\linewidth]{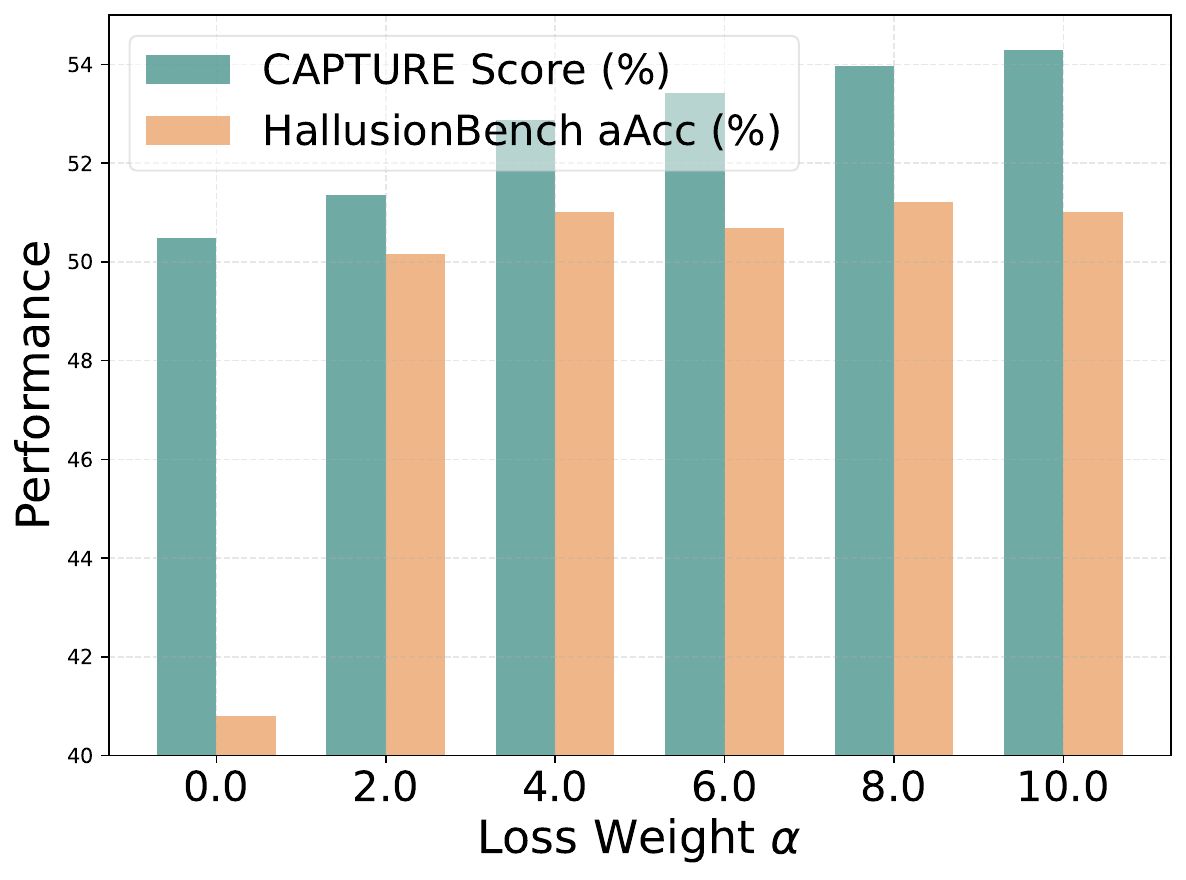}
    \caption{Loss Weight of Unlikelihood}
    \label{fig:abl-lossw}
\end{subcaptionblock}
\vspace{-3mm}
\caption{(a) Increase in activated average image attention across reflection turns. (b-d) Ablation studies.}
\label{fig:ablations}
\vspace{-4mm}
\end{figure*}

\paragraph{Reflection Turns} 
We analyze the impact of reflection turns on model performance using LLaVA-Critic and GPT-4o as the critic. As shown in~\cref{fig:abl-round}, increasing reflection turns improves performance on the DetailCaps-4870 benchmark, reducing hallucinations and enhancing detail perception. This aligns with our attention analysis (\cref{fig:attn_compare}), suggesting that iterative reflection helps the model better focus on relevant image regions.

\paragraph{Scoring Disparity for Data Construction} 
We also examine the effect of scoring thresholds in data selection (\cref{sec:rpl}) on DetailCaps and HallusionBench. As shown in~\cref{fig:abl-disparity}, optimal performance is achieved with samples having highest scores above 9 and score disparities of at least 4, indicating that high scoring disparity helps select challenging yet high-quality training samples.

\paragraph{Unlikelihood Loss} 
We further study the influence of unlikelihood loss weight $\alpha$ (from \cref{eq:policy}) on reducing behavior collapse in initial responses using DetailCaps and HallusionBench. As shown in~\cref{fig:abl-lossw}, a weight of 10.0 achieves optimal performance by effectively balancing the penalization of undesirable responses while preserving valuable content. 
\subsection{Further Analysis}
\label{exp_analysis}
\begin{table}
\centering
\caption{Comparison of RePer and RePer without RPL under varying critics and reflection turns on Detailcaps-4870.}
\label{tab:capture_scores}
\resizebox{0.9\linewidth}{!}{\begin{tabular}{l|c|cc}
\toprule
\textbf{Critic} & \textbf{Turn} & \textbf{RePer} & \textbf{RePer w.o. RPL} \\
\midrule
\multirow{3}{*}{GPT-4o~\citep{gpt4o}} & 1  &54.29 & 51.22 \\
                                & 2	&55.41 & 52.28 \\
                                & 3 &55.55 & 53.9 \\
                               
\midrule
\multirow{3}{*}{LLaVA-Critic~\citep{llavacritic}} & 1 &54.29 & 51.22 \\
                                    & 2 & 54.68 & 52.25 \\
                                    & 3 &54.73 & 53.85 \\
                                 
\bottomrule
\end{tabular}}
\vspace{-4mm}
\label{tb:abl-rounds}
\end{table}
\textbf{Critic matters, RPL matters more.}
To assess RPL and different critics’ impact on RePer, we compare its performance with and without RPL, using critics LLaVA-Critic~\citep{llavacritic} and GPT-4o~\citep{gpt4o}, across multiple reflection turns on DetailCaps. As shown in~\cref{tb:abl-rounds}, GPT-4o yields superior results due to its strong generative and discriminative abilities, while LLaVA-Critic also shows consistent improvements, indicating RePer’s adaptability to different critics. Even without RPL, RePer benefits from reflection; however, RPL further amplifies this effect, leading to a stronger initial-turn performance and demonstrating the effectiveness of the imitation learning approach.

\begin{table}
\centering
\caption{RPL vs. Preference Optimization Methods.}
\label{tb:exp-po}

\resizebox{0.47\textwidth}{!}{ 
\begin{tabular}{lccccc}
\toprule
\textbf{Method} & \textbf{DetailCaps}& \textbf{HallusionB} &  \textbf{GAIVE} & \textbf{LLaVAB}\\
\midrule
LLaVA-1.5-13B     & 51.22      & 24.43               &  5.65    & 66.95  \\
\midrule
+DPO~\citep{dpo}         & 50.53             & 25.61              &    5.28  & 66.2    \\
+LiPO~\citep{lipo}        &  52.31                 & 25.04         &   6.27   & \textbf{69.5}    \\
\rowcolor{mydred} +RPL& \textbf{54.73}    & \textbf{31.28}     &  \textbf{6.86}    & 67.6   \\
\bottomrule
\end{tabular}
}
\vspace{-3mm}
\end{table}

\textbf{RPL is essentially fine-grained preference optimization.} 
As detailed in \cref{sec:fpo}, RPL’s imitation learning in reflective dialogues can be seen as listwise preference optimization with detailed feedback and explicit rewards. We compare it to similar methods: DPO, which optimizes Bradley-Terry~\citep{btmodel} using preference pairs with the largest score differences, and LiPO, which optimizes learning-to-rank~\citep{learntorank} using all preference data ranked by reward. \cref{tb:exp-po} shows RPL’s clear advantages, especially in caption and hallucination metrics. We speculate this success stems from: 1) fine-grained critic feedback that facilitates effective corrections, lacking in DPO/LiPO; and 2) unlikelihood training without KL constraints, which helps counteract multimodal hallucinations.

\section{Related Work}
\label{sec:related_work}
The remarkable scaling laws~\citep{scalinglaw} of LLMs~\citep{llama1,xu2024restful} in terms of parameters and data have driven the advancement of LVLMs. BLIP-2~\citep{blip2} pioneered the use of Q-Former to bridge visual encoders with large language models, explicitly supervising the vision-language alignment while autoregressively generating vision-related text. Works like LLaVA~\citep{llava,llava1p5}, MiniGPT-4~\citep{minigpt4}, and Qwen-VL~\citep{qwenvl,qwen2vl} have demonstrated the sufficiency of text autoregression for visual understanding and have progressively simplified the vision-language connector using techniques such as cross-attention~\citep{mplug}, linear layers~\citep{llava,zhao2023chatspot}, MLPs~\citep{llava1p5,llavanext-video,dreamllm}, and convolutions~\citep{merlin,qwen2vl}, all while maintaining consistent performance.

Despite relentless scaling of visual encoders~\citep{cambrian,wei2024vary}, language decoders~\citep{qwen2vl}, and visual-textual corpora~\citep{li2024omnicorpus,wei2024general}, LVLMs have yet to achieve a qualitative leap in perceptual acuity or hallucination mitigation. Some approaches attribute hallucinations to visual~\citep{eyes} or linguistic biases~\citep{pope}, seeking to counter them through online~\citep{lrv} or offline~\citep{vcd} corrections. 
Others~\citep{rlhfv,seva,zhu2025perpo} take a more direct route, modulating the model’s visual attention preferences by aligning with human judgment via Reinforcement Learning from Human Feedback (RLHF)~\citep{instructgpt}. Yet, disappointingly, these efforts have failed to tackle the root issue: models still reflexively respond to perceptual challenges, regardless of their complexity.


LLMs often use step-by-step reasoning~\citep{cot} to avoid giving premature answers. However, this linear process can falter with complex problems, leading to factual inaccuracies and hallucinations~\citep{selfcheck}. To counter this, some approaches use external feedback to guide reasoning~\citep{reflexion,react}, while others harness the model’s reflective abilities for self-correction~\citep{selfcorrect,selfcheck,recursiveintro,score}. These methods employ an iterative “answer-reflect-reanswer” loop, significantly improving performance on complex challenges.


Some LVLMs require preliminary image parsing tasks like grounding~\citep{shikra,visualcot}, parsing~\citep{chartthinker,wei2024slow,chen2024onechart}, or identification~\citep{merlin, yu2025unhackable} before responding. While this chain-of-thought-style approach moderately improves performance, other methods~\citep{dualfocus,vstar} focus on locating relevant image regions and cropping them to assist with fine-grained perception. However, these methods often struggle with complex scenarios and may increase hallucination.
Recent work explores iterative refinement using internal~\citep{selfcorrect,volcano} or external~\citep{external} rewards. 
Despite promising results, these approaches lack systematic training frameworks and do not sufficiently explore the underlying principles of their mechanisms. We address these limitations by proposing RePer and RPL, with comprehensive theoretical and empirical analysis.
\section{Conclusion}

Perception in reflection addresses a key limitation in current LVLMs: the unrealistic expectation of perfect initial responses. Instead, it provides a robust fallback mechanism, empowering the model to adjust and converge on the correct answer even when initial predictions fall short. Powered by reflective perceptual learning, we create a system that can generalize more effectively across varied and complex visual scenarios, ensuring that the model is not only accurate but also resilient and adaptive in real-world applications.

\nocite{langley00}

\bibliography{main}

\begin{thebibliography}{80}
\providecommand{\natexlab}[1]{#1}
\providecommand{\url}[1]{\texttt{#1}}
\expandafter\ifx\csname urlstyle\endcsname\relax
  \providecommand{\doi}[1]{doi: #1}\else
  \providecommand{\doi}{doi: \begingroup \urlstyle{rm}\Url}\fi

\bibitem[Agrawal et~al.(2019)Agrawal, Desai, Wang, Chen, Jain, Johnson, Batra, Parikh, Lee, and Anderson]{agrawal2019nocaps}
Agrawal, H., Desai, K., Wang, Y., Chen, X., Jain, R., Johnson, M., Batra, D., Parikh, D., Lee, S., and Anderson, P.
\newblock Nocaps: Novel object captioning at scale.
\newblock In \emph{Proceedings of the IEEE/CVF international conference on computer vision}, pp.\  8948--8957, 2019.

\bibitem[Bai et~al.(2023)Bai, Bai, Yang, Wang, Tan, Wang, Lin, Zhou, and Zhou]{qwenvl}
Bai, J., Bai, S., Yang, S., Wang, S., Tan, S., Wang, P., Lin, J., Zhou, C., and Zhou, J.
\newblock Qwen-vl: A frontier large vision-language model with versatile abilities.
\newblock \emph{arXiv preprint arXiv:2308.12966}, 2023.

\bibitem[Betker et~al.(2023)Betker, Goh, Jing, Brooks, Wang, Li, Ouyang, Zhuang, Lee, Guo, et~al.]{dalle3}
Betker, J., Goh, G., Jing, L., Brooks, T., Wang, J., Li, L., Ouyang, L., Zhuang, J., Lee, J., Guo, Y., et~al.
\newblock Improving image generation with better captions.
\newblock \emph{Computer Science. https://cdn. openai. com/papers/dall-e-3. pdf}, 2\penalty0 (3):\penalty0 8, 2023.

\bibitem[Bradley \& Terry(1952)Bradley and Terry]{btmodel}
Bradley, R.~A. and Terry, M.~E.
\newblock Rank analysis of incomplete block designs: I. the method of paired comparisons.
\newblock \emph{Biometrika}, 39:\penalty0 324, 1952.
\newblock URL \url{https://api.semanticscholar.org/CorpusID:125209808}.

\bibitem[Cao et~al.(2024)Cao, Zhang, Dong, Lin, and Wang]{dualfocus}
Cao, Y., Zhang, P., Dong, X., Lin, D., and Wang, J.
\newblock Dualfocus: Integrating macro and micro perspectives in multi-modal large language models.
\newblock \emph{arXiv preprint arXiv:2402.14767}, 2024.

\bibitem[Caron et~al.(2021)Caron, Touvron, Misra, J\'egou, Mairal, Bojanowski, and Joulin]{caron2021emerging}
Caron, M., Touvron, H., Misra, I., J\'egou, H., Mairal, J., Bojanowski, P., and Joulin, A.
\newblock Emerging properties in self-supervised vision transformers.
\newblock In \emph{Proceedings of the International Conference on Computer Vision (ICCV)}, 2021.

\bibitem[Chen et~al.(2024)Chen, Kong, Wei, Liu, Ge, Zhao, Sun, Han, and Zhang]{chen2024onechart}
Chen, J., Kong, L., Wei, H., Liu, C., Ge, Z., Zhao, L., Sun, J., Han, C., and Zhang, X.
\newblock Onechart: Purify the chart structural extraction via one auxiliary token.
\newblock In \emph{Proceedings of the 32nd ACM International Conference on Multimedia}, pp.\  147--155, 2024.

\bibitem[Chen et~al.(2023)Chen, Zhang, Zeng, Zhang, Zhu, and Zhao]{shikra}
Chen, K., Zhang, Z., Zeng, W., Zhang, R., Zhu, F., and Zhao, R.
\newblock Shikra: Unleashing multimodal llm's referential dialogue magic.
\newblock \emph{arXiv preprint arXiv:2306.15195}, 2023.

\bibitem[Chen et~al.(2025)Chen, Li, Dong, Zhang, He, Wang, Zhao, and Lin]{chen2025sharegpt4v}
Chen, L., Li, J., Dong, X., Zhang, P., He, C., Wang, J., Zhao, F., and Lin, D.
\newblock Sharegpt4v: Improving large multi-modal models with better captions.
\newblock In \emph{European Conference on Computer Vision}, pp.\  370--387. Springer, 2025.

\bibitem[Chen et~al.(2015)Chen, Fang, Lin, Vedantam, Gupta, Doll{\'a}r, and Zitnick]{chen2015microsoft}
Chen, X., Fang, H., Lin, T.-Y., Vedantam, R., Gupta, S., Doll{\'a}r, P., and Zitnick, C.~L.
\newblock Microsoft coco captions: Data collection and evaluation server.
\newblock \emph{arXiv preprint arXiv:1504.00325}, 2015.

\bibitem[Dai et~al.(2024)Dai, Li, Li, Tiong, Zhao, Wang, Li, Fung, and Hoi]{instructblip}
Dai, W., Li, J., Li, D., Tiong, A. M.~H., Zhao, J., Wang, W., Li, B., Fung, P.~N., and Hoi, S.
\newblock Instructblip: Towards general-purpose vision-language models with instruction tuning.
\newblock \emph{Advances in Neural Information Processing Systems}, 36, 2024.

\bibitem[Darcet et~al.(2023)Darcet, Oquab, Mairal, and Bojanowski]{darcet2023vitneedreg}
Darcet, T., Oquab, M., Mairal, J., and Bojanowski, P.
\newblock Vision transformers need registers, 2023.

\bibitem[Devlin et~al.(2018)Devlin, Chang, Lee, and Toutanova]{bert}
Devlin, J., Chang, M.-W., Lee, K., and Toutanova, K.
\newblock Bert: Pre-training of deep bidirectional transformers for language understanding.
\newblock \emph{arXiv preprint arXiv:1810.04805}, 2018.

\bibitem[Dong et~al.(2024{\natexlab{a}})Dong, Li, Wu, Wang, Zhang, and Guo]{dong2024benchmarking}
Dong, H., Li, J., Wu, B., Wang, J., Zhang, Y., and Guo, H.
\newblock Benchmarking and improving detail image caption.
\newblock \emph{arXiv preprint arXiv:2405.19092}, 2024{\natexlab{a}}.

\bibitem[Dong et~al.(2024{\natexlab{b}})Dong, Han, Peng, Qi, Ge, Yang, Zhao, Sun, Zhou, Wei, Kong, Zhang, Ma, and Yi]{dreamllm}
Dong, R., Han, C., Peng, Y., Qi, Z., Ge, Z., Yang, J., Zhao, L., Sun, J., Zhou, H., Wei, H., Kong, X., Zhang, X., Ma, K., and Yi, L.
\newblock Dream{LLM}: Synergistic multimodal comprehension and creation.
\newblock In \emph{The Twelfth International Conference on Learning Representations}, 2024{\natexlab{b}}.

\bibitem[Ge et~al.(2024)Ge, Zeng, Huffman, Lin, Liu, and Cui]{ge2024visual}
Ge, Y., Zeng, X., Huffman, J.~S., Lin, T.-Y., Liu, M.-Y., and Cui, Y.
\newblock Visual fact checker: Enabling high-fidelity detailed caption generation.
\newblock In \emph{Proceedings of the IEEE/CVF Conference on Computer Vision and Pattern Recognition}, pp.\  14033--14042, 2024.

\bibitem[GPT-4o(2024)]{gpt4o}
GPT-4o.
\newblock Hello gpt-4o, 2024.
\newblock URL \url{https://openai.com/index/hello-gpt-4o/}.

\bibitem[Guan et~al.(2024)Guan, Liu, Wu, Xian, Li, Liu, Wang, Chen, Huang, Yacoob, et~al.]{guan2024hallusionbench}
Guan, T., Liu, F., Wu, X., Xian, R., Li, Z., Liu, X., Wang, X., Chen, L., Huang, F., Yacoob, Y., et~al.
\newblock Hallusionbench: an advanced diagnostic suite for entangled language hallucination and visual illusion in large vision-language models.
\newblock In \emph{Proceedings of the IEEE/CVF Conference on Computer Vision and Pattern Recognition}, pp.\  14375--14385, 2024.

\bibitem[He et~al.(2016)He, Zhang, Ren, and Sun]{resnet}
He, K., Zhang, X., Ren, S., and Sun, J.
\newblock Deep residual learning for image recognition.
\newblock In \emph{Proceedings of the IEEE conference on computer vision and pattern recognition}, pp.\  770--778, 2016.

\bibitem[He et~al.(2017)He, Gkioxari, Doll{\'a}r, and Girshick]{maskrcnn}
He, K., Gkioxari, G., Doll{\'a}r, P., and Girshick, R.
\newblock Mask r-cnn.
\newblock In \emph{Proceedings of the IEEE international conference on computer vision}, pp.\  2961--2969, 2017.

\bibitem[Huang et~al.(2024)Huang, Dong, Zhang, Wang, He, Wang, Lin, Zhang, and Yu]{opera}
Huang, Q., Dong, X., Zhang, P., Wang, B., He, C., Wang, J., Lin, D., Zhang, W., and Yu, N.
\newblock Opera: Alleviating hallucination in multi-modal large language models via over-trust penalty and retrospection-allocation.
\newblock In \emph{Proceedings of the IEEE/CVF Conference on Computer Vision and Pattern Recognition}, pp.\  13418--13427, 2024.

\bibitem[Kaplan et~al.(2020)Kaplan, McCandlish, Henighan, Brown, Chess, Child, Gray, Radford, Wu, and Amodei]{scalinglaw}
Kaplan, J., McCandlish, S., Henighan, T., Brown, T.~B., Chess, B., Child, R., Gray, S., Radford, A., Wu, J., and Amodei, D.
\newblock Scaling laws for neural language models.
\newblock \emph{arXiv preprint arXiv:2001.08361}, 2020.

\bibitem[Kim et~al.(2024)Kim, Pertsch, Karamcheti, Xiao, Balakrishna, Nair, Rafailov, Foster, Lam, Sanketi, et~al.]{vla2}
Kim, M.~J., Pertsch, K., Karamcheti, S., Xiao, T., Balakrishna, A., Nair, S., Rafailov, R., Foster, E., Lam, G., Sanketi, P., et~al.
\newblock Openvla: An open-source vision-language-action model.
\newblock \emph{arXiv preprint arXiv:2406.09246}, 2024.

\bibitem[Kumar et~al.(2024)Kumar, Zhuang, Agarwal, Su, Co-Reyes, Singh, Baumli, Iqbal, Bishop, Roelofs, et~al.]{score}
Kumar, A., Zhuang, V., Agarwal, R., Su, Y., Co-Reyes, J.~D., Singh, A., Baumli, K., Iqbal, S., Bishop, C., Roelofs, R., et~al.
\newblock Training language models to self-correct via reinforcement learning.
\newblock \emph{arXiv preprint arXiv:2409.12917}, 2024.

\bibitem[Lee et~al.(2023)Lee, Park, Jo, and Seo]{volcano}
Lee, S., Park, S.~H., Jo, Y., and Seo, M.
\newblock Volcano: mitigating multimodal hallucination through self-feedback guided revision.
\newblock \emph{arXiv preprint arXiv:2311.07362}, 2023.

\bibitem[Leng et~al.(2024)Leng, Zhang, Chen, Li, Lu, Miao, and Bing]{vcd}
Leng, S., Zhang, H., Chen, G., Li, X., Lu, S., Miao, C., and Bing, L.
\newblock Mitigating object hallucinations in large vision-language models through visual contrastive decoding.
\newblock In \emph{Proceedings of the IEEE/CVF Conference on Computer Vision and Pattern Recognition}, pp.\  13872--13882, 2024.

\bibitem[Li et~al.(2023{\natexlab{a}})Li, Li, Savarese, and Hoi]{blip2}
Li, J., Li, D., Savarese, S., and Hoi, S.
\newblock Blip-2: Bootstrapping language-image pre-training with frozen image encoders and large language models.
\newblock In \emph{International conference on machine learning}, pp.\  19730--19742. PMLR, 2023{\natexlab{a}}.

\bibitem[Li et~al.(2024)Li, Chen, Wang, Wang, Ye, Jin, Chen, He, Gao, Cui, et~al.]{li2024omnicorpus}
Li, Q., Chen, Z., Wang, W., Wang, W., Ye, S., Jin, Z., Chen, G., He, Y., Gao, Z., Cui, E., et~al.
\newblock Omnicorpus: An unified multimodal corpus of 10 billion-level images interleaved with text.
\newblock \emph{arXiv preprint arXiv:2406.08418}, 2024.

\bibitem[Li et~al.(2023{\natexlab{b}})Li, Du, Zhou, Wang, Zhao, and Wen]{pope}
Li, Y., Du, Y., Zhou, K., Wang, J., Zhao, W.~X., and Wen, J.-R.
\newblock Evaluating object hallucination in large vision-language models.
\newblock \emph{arXiv preprint arXiv:2305.10355}, 2023{\natexlab{b}}.

\bibitem[Li et~al.(2023{\natexlab{c}})Li, Chai, Zhuo, Qu, Haffari, Li, Ji, and Tran]{li2023factual}
Li, Z., Chai, Y., Zhuo, T.~Y., Qu, L., Haffari, G., Li, F., Ji, D., and Tran, Q.~H.
\newblock Factual: A benchmark for faithful and consistent textual scene graph parsing.
\newblock \emph{arXiv preprint arXiv:2305.17497}, 2023{\natexlab{c}}.

\bibitem[Liao et~al.(2024)Liao, Mahmood, Fidler, and Acuna]{external}
Liao, Y.-H., Mahmood, R., Fidler, S., and Acuna, D.
\newblock Can feedback enhance semantic grounding in large vision-language models?
\newblock \emph{arXiv preprint arXiv:2404.06510}, 2024.

\bibitem[Liu et~al.(2023{\natexlab{a}})Liu, Lin, Li, Wang, Yacoob, and Wang]{liu2023aligning}
Liu, F., Lin, K., Li, L., Wang, J., Yacoob, Y., and Wang, L.
\newblock Aligning large multi-modal model with robust instruction tuning.
\newblock \emph{arXiv preprint arXiv:2306.14565}, 2023{\natexlab{a}}.

\bibitem[Liu et~al.(2023{\natexlab{b}})Liu, Lin, Li, Wang, Yacoob, and Wang]{lrv}
Liu, F., Lin, K., Li, L., Wang, J., Yacoob, Y., and Wang, L.
\newblock Mitigating hallucination in large multi-modal models via robust instruction tuning.
\newblock In \emph{The Twelfth International Conference on Learning Representations}, 2023{\natexlab{b}}.

\bibitem[Liu et~al.(2024{\natexlab{a}})Liu, Mao, Cao, Xue, Johnson, Tang, and Wang]{selfcorrect}
Liu, G., Mao, H., Cao, B., Xue, Z., Johnson, K., Tang, J., and Wang, R.
\newblock On the intrinsic self-correction capability of llms: Uncertainty and latent concept.
\newblock \emph{arXiv preprint arXiv:2406.02378}, 2024{\natexlab{a}}.

\bibitem[Liu et~al.(2024{\natexlab{b}})Liu, Li, Li, and Lee]{llava1p5}
Liu, H., Li, C., Li, Y., and Lee, Y.~J.
\newblock Improved baselines with visual instruction tuning.
\newblock In \emph{Proceedings of the IEEE/CVF Conference on Computer Vision and Pattern Recognition}, pp.\  26296--26306, 2024{\natexlab{b}}.

\bibitem[Liu et~al.(2024{\natexlab{c}})Liu, Li, Wu, and Lee]{llava}
Liu, H., Li, C., Wu, Q., and Lee, Y.~J.
\newblock Visual instruction tuning.
\newblock \emph{Advances in neural information processing systems}, 36, 2024{\natexlab{c}}.

\bibitem[Liu et~al.(2024{\natexlab{d}})Liu, Chen, Li, Fang, and Shen]{chartthinker}
Liu, M., Chen, D., Li, Y., Fang, G., and Shen, Y.
\newblock Chartthinker: A contextual chain-of-thought approach to optimized chart summarization.
\newblock \emph{arXiv preprint arXiv:2403.11236}, 2024{\natexlab{d}}.

\bibitem[Liu et~al.(2024{\natexlab{e}})Liu, Qin, Wu, Shen, Khalman, Joshi, Zhao, Saleh, Baumgartner, Liu, et~al.]{lipo}
Liu, T., Qin, Z., Wu, J., Shen, J., Khalman, M., Joshi, R., Zhao, Y., Saleh, M., Baumgartner, S., Liu, J., et~al.
\newblock Lipo: Listwise preference optimization through learning-to-rank.
\newblock \emph{arXiv preprint arXiv:2402.01878}, 2024{\natexlab{e}}.

\bibitem[Liu et~al.(2009)]{learntorank}
Liu, T.-Y. et~al.
\newblock Learning to rank for information retrieval.
\newblock \emph{Foundations and Trends{\textregistered} in Information Retrieval}, 3\penalty0 (3):\penalty0 225--331, 2009.

\bibitem[Ma{\l}ki{\'n}ski \& Ma{\'n}dziuk(2022)Ma{\l}ki{\'n}ski and Ma{\'n}dziuk]{visualreason}
Ma{\l}ki{\'n}ski, M. and Ma{\'n}dziuk, J.
\newblock Deep learning methods for abstract visual reasoning: A survey on raven's progressive matrices.
\newblock \emph{ACM Computing Surveys}, 2022.

\bibitem[Miao et~al.(2023)Miao, Teh, and Rainforth]{selfcheck}
Miao, N., Teh, Y.~W., and Rainforth, T.
\newblock Selfcheck: Using llms to zero-shot check their own step-by-step reasoning.
\newblock \emph{arXiv preprint arXiv:2308.00436}, 2023.

\bibitem[Miller(1995)]{miller1995wordnet}
Miller, G.~A.
\newblock Wordnet: a lexical database for english.
\newblock \emph{Communications of the ACM}, 38\penalty0 (11):\penalty0 39--41, 1995.

\bibitem[Mu et~al.(2024)Mu, Helyar, Heidecke, Achiam, Vallone, Kivlichan, Lin, Beutel, Schulman, and Weng]{mu2024rule}
Mu, T., Helyar, A., Heidecke, J., Achiam, J., Vallone, A., Kivlichan, I., Lin, M., Beutel, A., Schulman, J., and Weng, L.
\newblock Rule based rewards for language model safety.
\newblock \emph{arXiv preprint arXiv:2411.01111}, 2024.

\bibitem[Ouyang et~al.(2022)Ouyang, Wu, Jiang, Almeida, Wainwright, Mishkin, Zhang, Agarwal, Slama, Ray, et~al.]{instructgpt}
Ouyang, L., Wu, J., Jiang, X., Almeida, D., Wainwright, C., Mishkin, P., Zhang, C., Agarwal, S., Slama, K., Ray, A., et~al.
\newblock Training language models to follow instructions with human feedback.
\newblock \emph{Advances in neural information processing systems}, 35:\penalty0 27730--27744, 2022.

\bibitem[Peng et~al.(2024)Peng, Cui, Tang, Qi, Dong, Bai, Han, Ge, Zhang, and Xia]{peng2024dreambench++}
Peng, Y., Cui, Y., Tang, H., Qi, Z., Dong, R., Bai, J., Han, C., Ge, Z., Zhang, X., and Xia, S.-T.
\newblock Dreambench++: A human-aligned benchmark for personalized image generation.
\newblock \emph{arXiv preprint arXiv:2406.16855}, 2024.

\bibitem[Qu et~al.(2024)Qu, Zhang, Garg, and Kumar]{recursiveintro}
Qu, Y., Zhang, T., Garg, N., and Kumar, A.
\newblock Recursive introspection: Teaching language model agents how to self-improve.
\newblock \emph{arXiv preprint arXiv:2407.18219}, 2024.

\bibitem[Rafailov et~al.(2024)Rafailov, Sharma, Mitchell, Manning, Ermon, and Finn]{dpo}
Rafailov, R., Sharma, A., Mitchell, E., Manning, C.~D., Ermon, S., and Finn, C.
\newblock Direct preference optimization: Your language model is secretly a reward model.
\newblock \emph{Advances in Neural Information Processing Systems}, 36, 2024.

\bibitem[Ren et~al.(2016)Ren, He, Girshick, and Sun]{fasterrcnn}
Ren, S., He, K., Girshick, R., and Sun, J.
\newblock Faster r-cnn: Towards real-time object detection with region proposal networks.
\newblock \emph{IEEE transactions on pattern analysis and machine intelligence}, 39\penalty0 (6):\penalty0 1137--1149, 2016.

\bibitem[Ross et~al.(2011)Ross, Gordon, and Bagnell]{imitation}
Ross, S., Gordon, G., and Bagnell, D.
\newblock A reduction of imitation learning and structured prediction to no-regret online learning.
\newblock In \emph{Proceedings of the fourteenth international conference on artificial intelligence and statistics}, pp.\  627--635. JMLR Workshop and Conference Proceedings, 2011.

\bibitem[Sampat et~al.(2022)Sampat, Patel, Das, Yang, and Baral]{vla1}
Sampat, S.~K., Patel, M., Das, S., Yang, Y., and Baral, C.
\newblock Reasoning about actions over visual and linguistic modalities: A survey.
\newblock \emph{arXiv preprint arXiv:2207.07568}, 2022.

\bibitem[Shao et~al.(2024)Shao, Qian, Xiao, Song, Zong, Wang, Liu, and Li]{visualcot}
Shao, H., Qian, S., Xiao, H., Song, G., Zong, Z., Wang, L., Liu, Y., and Li, H.
\newblock Visual cot: Unleashing chain-of-thought reasoning in multi-modal language models.
\newblock \emph{arXiv preprint arXiv:2403.16999}, 2024.

\bibitem[Shinn et~al.(2024)Shinn, Cassano, Gopinath, Narasimhan, and Yao]{reflexion}
Shinn, N., Cassano, F., Gopinath, A., Narasimhan, K., and Yao, S.
\newblock Reflexion: Language agents with verbal reinforcement learning.
\newblock \emph{Advances in Neural Information Processing Systems}, 36, 2024.

\bibitem[Sun et~al.(2023{\natexlab{a}})Sun, Shen, Cao, Liu, Li, Shen, Gan, Gui, Wang, Yang, Keutzer, and Darrell]{2023llavarlhf}
Sun, Z., Shen, S., Cao, S., Liu, H., Li, C., Shen, Y., Gan, C., Gui, L.-Y., Wang, Y.-X., Yang, Y., Keutzer, K., and Darrell, T.
\newblock Aligning large multimodal models with factually augmented rlhf.
\newblock 2023{\natexlab{a}}.

\bibitem[Sun et~al.(2023{\natexlab{b}})Sun, Shen, Cao, Liu, Li, Shen, Gan, Gui, Wang, Yang, Keutzer, and Darrell]{sun2023aligninglargemultimodalmodels}
Sun, Z., Shen, S., Cao, S., Liu, H., Li, C., Shen, Y., Gan, C., Gui, L.-Y., Wang, Y.-X., Yang, Y., Keutzer, K., and Darrell, T.
\newblock Aligning large multimodal models with factually augmented rlhf, 2023{\natexlab{b}}.
\newblock URL \url{https://arxiv.org/abs/2309.14525}.

\bibitem[Swamy et~al.(2023)Swamy, Wu, Choudhury, Bagnell, and Wu]{swamy2023inverse}
Swamy, G., Wu, D., Choudhury, S., Bagnell, D., and Wu, S.
\newblock Inverse reinforcement learning without reinforcement learning.
\newblock In \emph{International Conference on Machine Learning}, pp.\  33299--33318. PMLR, 2023.

\bibitem[Tang et~al.(2024)Tang, Guo, Zheng, Calandriello, Cao, Tarassov, Munos, Pires, Valko, Cheng, et~al.]{online}
Tang, Y., Guo, D.~Z., Zheng, Z., Calandriello, D., Cao, Y., Tarassov, E., Munos, R., Pires, B.~{\'A}., Valko, M., Cheng, Y., et~al.
\newblock Understanding the performance gap between online and offline alignment algorithms.
\newblock \emph{arXiv preprint arXiv:2405.08448}, 2024.

\bibitem[Tong et~al.(2024{\natexlab{a}})Tong, Brown, Wu, Woo, Middepogu, Akula, Yang, Yang, Iyer, Pan, et~al.]{cambrian}
Tong, S., Brown, E., Wu, P., Woo, S., Middepogu, M., Akula, S.~C., Yang, J., Yang, S., Iyer, A., Pan, X., et~al.
\newblock Cambrian-1: A fully open, vision-centric exploration of multimodal llms.
\newblock \emph{arXiv preprint arXiv:2406.16860}, 2024{\natexlab{a}}.

\bibitem[Tong et~al.(2024{\natexlab{b}})Tong, Liu, Zhai, Ma, LeCun, and Xie]{eyes}
Tong, S., Liu, Z., Zhai, Y., Ma, Y., LeCun, Y., and Xie, S.
\newblock Eyes wide shut? exploring the visual shortcomings of multimodal llms.
\newblock In \emph{Proceedings of the IEEE/CVF Conference on Computer Vision and Pattern Recognition}, pp.\  9568--9578, 2024{\natexlab{b}}.

\bibitem[Touvron et~al.(2023{\natexlab{a}})Touvron, Lavril, Izacard, Martinet, Lachaux, Lacroix, Rozi{\`e}re, Goyal, Hambro, Azhar, et~al.]{llama1}
Touvron, H., Lavril, T., Izacard, G., Martinet, X., Lachaux, M.-A., Lacroix, T., Rozi{\`e}re, B., Goyal, N., Hambro, E., Azhar, F., et~al.
\newblock Llama: Open and efficient foundation language models.
\newblock \emph{arXiv preprint arXiv:2302.13971}, 2023{\natexlab{a}}.

\bibitem[Touvron et~al.(2023{\natexlab{b}})Touvron, Martin, Stone, Albert, Almahairi, Babaei, Bashlykov, Batra, Bhargava, Bhosale, et~al.]{touvron2023llama}
Touvron, H., Martin, L., Stone, K., Albert, P., Almahairi, A., Babaei, Y., Bashlykov, N., Batra, S., Bhargava, P., Bhosale, S., et~al.
\newblock Llama 2: Open foundation and fine-tuned chat models.
\newblock \emph{arXiv preprint arXiv:2307.09288}, 2023{\natexlab{b}}.

\bibitem[Wang et~al.(2024)Wang, Bai, Tan, Wang, Fan, Bai, Chen, Liu, Wang, Ge, et~al.]{qwen2vl}
Wang, P., Bai, S., Tan, S., Wang, S., Fan, Z., Bai, J., Chen, K., Liu, X., Wang, J., Ge, W., et~al.
\newblock Qwen2-vl: Enhancing vision-language model's perception of the world at any resolution.
\newblock \emph{arXiv preprint arXiv:2409.12191}, 2024.

\bibitem[Wei et~al.(2024{\natexlab{a}})Wei, Kong, Chen, Zhao, Ge, Yang, Sun, Han, and Zhang]{wei2024vary}
Wei, H., Kong, L., Chen, J., Zhao, L., Ge, Z., Yang, J., Sun, J., Han, C., and Zhang, X.
\newblock Vary: Scaling up the vision vocabulary for large vision-language model.
\newblock In \emph{European Conference on Computer Vision}, pp.\  408--424. Springer, 2024{\natexlab{a}}.

\bibitem[Wei et~al.(2024{\natexlab{b}})Wei, Liu, Chen, Wang, Kong, Xu, Ge, Zhao, Sun, Peng, et~al.]{wei2024general}
Wei, H., Liu, C., Chen, J., Wang, J., Kong, L., Xu, Y., Ge, Z., Zhao, L., Sun, J., Peng, Y., et~al.
\newblock General ocr theory: Towards ocr-2.0 via a unified end-to-end model.
\newblock 2024{\natexlab{b}}.

\bibitem[Wei et~al.(2024{\natexlab{c}})Wei, Yin, Li, Wang, Zhao, Sun, Ge, and Zhang]{wei2024slow}
Wei, H., Yin, Y., Li, Y., Wang, J., Zhao, L., Sun, J., Ge, Z., and Zhang, X.
\newblock Slow perception: Let's perceive geometric figures step-by-step.
\newblock \emph{arXiv preprint arXiv:2412.20631}, 2024{\natexlab{c}}.

\bibitem[Wei et~al.(2022)Wei, Wang, Schuurmans, Bosma, Xia, Chi, Le, Zhou, et~al.]{cot}
Wei, J., Wang, X., Schuurmans, D., Bosma, M., Xia, F., Chi, E., Le, Q.~V., Zhou, D., et~al.
\newblock Chain-of-thought prompting elicits reasoning in large language models.
\newblock \emph{Advances in neural information processing systems}, 35:\penalty0 24824--24837, 2022.

\bibitem[Welleck et~al.(2019)Welleck, Kulikov, Roller, Dinan, Cho, and Weston]{unlikelihood}
Welleck, S., Kulikov, I., Roller, S., Dinan, E., Cho, K., and Weston, J.
\newblock Neural text generation with unlikelihood training.
\newblock \emph{arXiv preprint arXiv:1908.04319}, 2019.

\bibitem[Wu \& Xie(2024)Wu and Xie]{vstar}
Wu, P. and Xie, S.
\newblock V?: Guided visual search as a core mechanism in multimodal llms.
\newblock In \emph{Proceedings of the IEEE/CVF Conference on Computer Vision and Pattern Recognition}, pp.\  13084--13094, 2024.

\bibitem[Xie et~al.(2024)Xie, Chen, Zhang, Wan, and Li]{agent1}
Xie, J., Chen, Z., Zhang, R., Wan, X., and Li, G.
\newblock Large multimodal agents: A survey.
\newblock \emph{arXiv preprint arXiv:2402.15116}, 2024.

\bibitem[Xiong et~al.(2024)Xiong, Wang, Guo, Ye, Fan, Gu, Huang, and Li]{llavacritic}
Xiong, T., Wang, X., Guo, D., Ye, Q., Fan, H., Gu, Q., Huang, H., and Li, C.
\newblock Llava-critic: Learning to evaluate multimodal models.
\newblock \emph{arXiv preprint arXiv:2410.02712}, 2024.

\bibitem[Xu et~al.(2024)Xu, Zhao, Wang, and Chen]{xu2024restful}
Xu, H., Zhao, R., Wang, J., and Chen, H.
\newblock Restful-llama: Connecting user queries to restful apis.
\newblock In \emph{Proceedings of the 2024 Conference on Empirical Methods in Natural Language Processing: Industry Track}, pp.\  1433--1443, 2024.

\bibitem[Yao et~al.(2022)Yao, Zhao, Yu, Du, Shafran, Narasimhan, and Cao]{react}
Yao, S., Zhao, J., Yu, D., Du, N., Shafran, I., Narasimhan, K., and Cao, Y.
\newblock React: Synergizing reasoning and acting in language models.
\newblock \emph{arXiv preprint arXiv:2210.03629}, 2022.

\bibitem[Ye et~al.(2023)Ye, Xu, Xu, Ye, Yan, Zhou, Wang, Hu, Shi, Shi, et~al.]{mplug}
Ye, Q., Xu, H., Xu, G., Ye, J., Yan, M., Zhou, Y., Wang, J., Hu, A., Shi, P., Shi, Y., et~al.
\newblock mplug-owl: Modularization empowers large language models with multimodality.
\newblock \emph{arXiv preprint arXiv:2304.14178}, 2023.

\bibitem[Yu et~al.(2023)Yu, Zhao, Wei, Yang, Wu, Kong, Wei, Wang, Ge, Zhang, et~al.]{merlin}
Yu, E., Zhao, L., Wei, Y., Yang, J., Wu, D., Kong, L., Wei, H., Wang, T., Ge, Z., Zhang, X., et~al.
\newblock Merlin: Empowering multimodal llms with foresight minds.
\newblock \emph{arXiv preprint arXiv:2312.00589}, 2023.

\bibitem[Yu et~al.(2025)Yu, Lin, Zhao, Wei, Zhu, Wei, Sun, Ge, Zhang, Wang, et~al.]{yu2025unhackable}
Yu, E., Lin, K., Zhao, L., Wei, Y., Zhu, Z., Wei, H., Sun, J., Ge, Z., Zhang, X., Wang, J., et~al.
\newblock Unhackable temporal rewarding for scalable video mllms.
\newblock \emph{arXiv preprint arXiv:2502.12081}, 2025.

\bibitem[Yu et~al.(2024)Yu, Yao, Zhang, He, Han, Cui, Hu, Liu, Zheng, Sun, et~al.]{rlhfv}
Yu, T., Yao, Y., Zhang, H., He, T., Han, Y., Cui, G., Hu, J., Liu, Z., Zheng, H.-T., Sun, M., et~al.
\newblock Rlhf-v: Towards trustworthy mllms via behavior alignment from fine-grained correctional human feedback.
\newblock In \emph{Proceedings of the IEEE/CVF Conference on Computer Vision and Pattern Recognition}, pp.\  13807--13816, 2024.

\bibitem[Zhang et~al.(2024)Zhang, Li, Liu, Lee, Gui, Fu, Feng, Liu, and Li]{llavanext-video}
Zhang, Y., Li, B., Liu, h., Lee, Y.~j., Gui, L., Fu, D., Feng, J., Liu, Z., and Li, C.
\newblock Llava-next: A strong zero-shot video understanding model, April 2024.
\newblock URL \url{https://llava-vl.github.io/blog/2024-04-30-llava-next-video/}.

\bibitem[Zhao et~al.(2023)Zhao, Yu, Ge, Yang, Wei, Zhou, Sun, Peng, Dong, Han, et~al.]{zhao2023chatspot}
Zhao, L., Yu, E., Ge, Z., Yang, J., Wei, H., Zhou, H., Sun, J., Peng, Y., Dong, R., Han, C., et~al.
\newblock Chatspot: Bootstrapping multimodal llms via precise referring instruction tuning.
\newblock \emph{arXiv preprint arXiv:2307.09474}, 2023.

\bibitem[Zhu et~al.(2023)Zhu, Chen, Shen, Li, and Elhoseiny]{minigpt4}
Zhu, D., Chen, J., Shen, X., Li, X., and Elhoseiny, M.
\newblock Minigpt-4: Enhancing vision-language understanding with advanced large language models, 2023.

\bibitem[Zhu et~al.(2024)Zhu, Zhao, Ge, and Zhang]{seva}
Zhu, K., Zhao, L., Ge, Z., and Zhang, X.
\newblock Self-supervised visual preference alignment.
\newblock In \emph{Proceedings of the 32nd ACM International Conference on Multimedia}, pp.\  291--300, 2024.

\bibitem[Zhu et~al.(2025)Zhu, Zhao, Lin, Yang, Yu, Liu, Wei, Sun, Ge, and Zhang]{zhu2025perpo}
Zhu, Z., Zhao, L., Lin, K., Yang, J., Yu, E., Liu, C., Wei, H., Sun, J., Ge, Z., and Zhang, X.
\newblock Perpo: Perceptual preference optimization via discriminative rewarding.
\newblock \emph{arXiv preprint arXiv:2502.04371}, 2025.

\end{thebibliography}
\bibliographystyle{icml2025}

\newpage
\appendix
\onecolumn
\onecolumn
\section*{Appendix}
In this appendix, we provide additional details to complement the main paper. Specifically, \cref{sup:dataset} elaborates on the Visual Reflection Dataset, while \cref{sup:gape} presents details of the proposed GAPE benchmark along with its complete results. Finally, \cref{sup:case} showcases additional examples illustrating the strong capabilities of RePer.

\section{Construction Details of Visual Reflection Dataset}
\label{sup:dataset}
This section provides additional details on the data construction process introduced in~\cref{sec:rpl} and~\cref{exp_imp}.

\subsection{Step-1: Initial Candidate Generation}
To generate diverse responses, we employ \textit{temperature sampling}, producing eight candidate captions per image across different temperature values, ranging from $0.0$ to $1.4$ in increments of $0.2$. Higher temperatures generally lead to lower response quality, often introducing hallucinated objects or less precise descriptions.

\subsection{Step-2: VLM-Based Reward Scoring}
\begin{figure*}[t]
\centering
\includegraphics[width=1.0\linewidth]{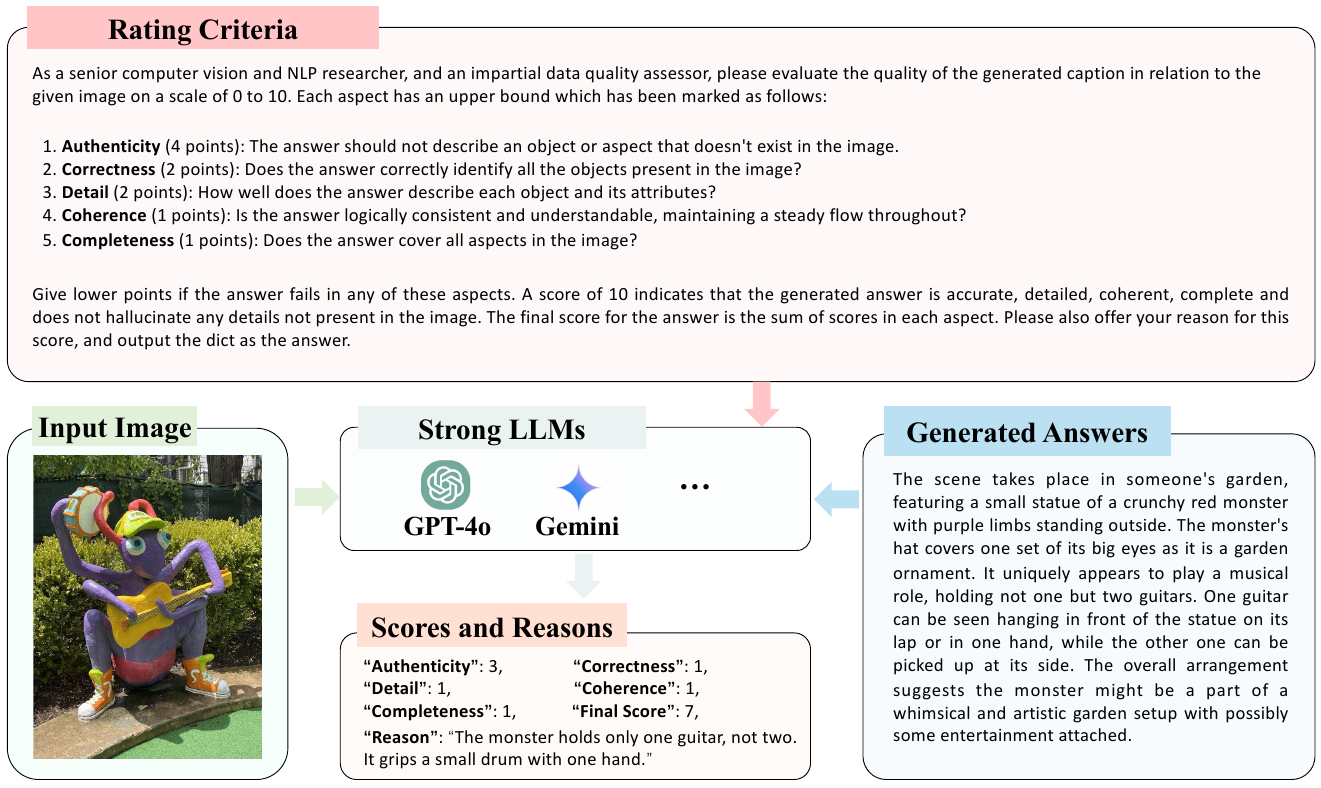}
\caption{\textbf{Pipeline for Generating VLM-Based Reward Scores.} Given an image and a generated answer, a predefined rating criterion for high-quality captions is formulated as a prompt. A strong LLM then evaluates the response, producing a reward score along with a rationale for the assigned score.}
\label{fig:data-con-case}
\end{figure*}

We define evaluation criteria for high-quality image captions, which guide the reward scoring process through carefully designed prompts (as shown in~\cref{fig:data-con-case}). The reward score ranges from 0 to 10 and assesses five key aspects:
\begin{itemize}
    \item Authenticity: Whether the caption contains hallucinated objects.
	\item Correctness: Whether all described attributes and relationships are factually correct.
	\item Detailness: Whether the description is sufficiently detailed, covering all relevant attributes of objects.
	\item Coherence: Whether the caption is logically consistent, without contradictions.
	\item Completeness: Whether the caption comprehensively covers all relevant aspects of the image, including both foreground and background elements.
\end{itemize}

\begin{figure*}[t]
\centering
\includegraphics[width=1.0\linewidth]{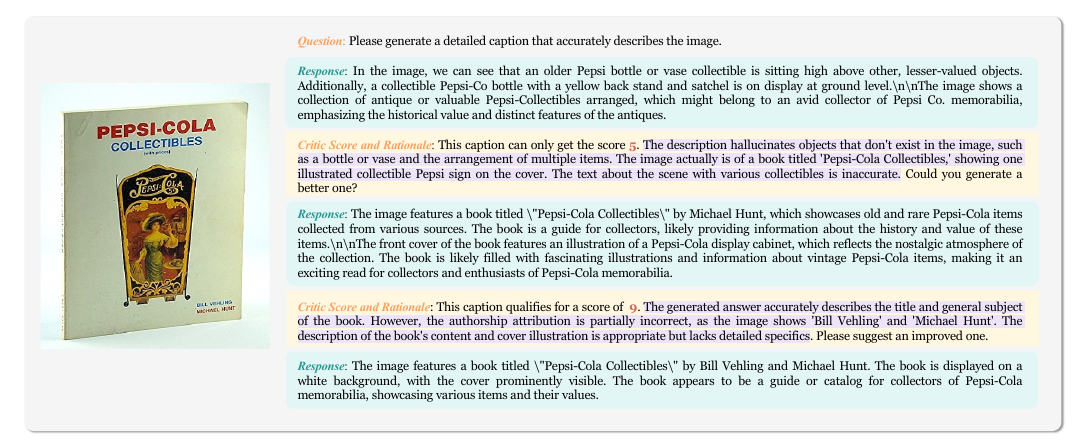}
\caption{\textbf{Illustration of data construction based on the designed template.} The \textcolor{responsecolor}{\textbf{responses}} are structured progressively from the lowest to the highest score. Each \textcolor{promptcolor}{\textbf{reflective prompt}} consists of three components: \textcolor{rewardcolor}{\textbf{reward score}}, \textcolor{reasoncolor}{\textbf{rationale}} for the score, and a follow-up question requesting an improved response. This iterative refinement process guides the model towards generating more accurate and detailed captions.}
\label{fig:data-reward-pipe}
\end{figure*}

\subsection{Step-3: Rule-Based Reward Scoring.}
Inspired by~\cite{dong2024benchmarking}, we design rule-based rewards to quantify the alignment between image elements and textual descriptions. This evaluates visual preference through a structured pipeline:
\paragraph{Reference Caption Generation}
We prompt strong VLMs (GPT-4o and Gemini-Pro) using \textit{“Please describe this image in detail.”} to generate reference captions for each image.
\paragraph{Element Extraction}
We extract objects, attributes, and relations from both reference captions and candidate answers using Factual Parser~\cite{li2023factual}, while applying stop-word filtering to remove irrelevant terms.
To filter irrelevant elements, a stop word list is curated for abstract nouns (e.g., “foreground”, “background”) that do not correspond to image content. LLaMA2-13B-chat~\cite{touvron2023llama} and Factual Parser are used to extract candidate nouns from ShareGPT4V-102k~\cite{chen2025sharegpt4v}. Words recalled by Factual Parser but missing in LLaMA2-13B-chat are reviewed, and high-frequency terms are validated by human experts. This process results in the final stop word list.

\paragraph{Elements Matching}
We implement a three-stage matching strategy to evaluate visual elements:
\begin{itemize}
    \item Exact Matching: Directly aligns identical objects, attributes, and relations.
    \item Synonym Matching: Uses WordNet~\cite{miller1995wordnet} to identify synonym sets and assigns a 1.0 match score for synonymous elements.
    \item Soft Matching: Applies BERT~\cite{bert} to compute cosine similarity between embeddings of unmatched elements, selecting the highest similarity score per element.
\end{itemize}
\paragraph{Final Matching Score}
The final Rule-based reward aggregates scores from all three stages to compute precision, recall, and F1-score. The final caption quality score is a weighted sum of the three F1 scores, with default weights of Object:Attribute:Relation = 5:2:2.

\subsection{Step-4: Reflective Dialogue Construction}
To construct reflective dialogues, we first apply data filtering based on the criteria outlined in \cref{sec:rpl} (Step-4) and \cref{exp_imp} (Datasets). We then organize responses into a structured reflection-driven dialogue format, using the template illustrated in \cref{fig:data-reward-pipe}.

For VLM-based rewards, given an image, its candidate answers, and scores with reasons from GPT-4o, we sort candidate answers from lowest to highest score. The multi-turn dialogue is constructed as follows:
\begin{itemize}
    \item In the first turn, the model is given the initial question.
	\item In subsequent turns, each reflective prompt includes the score, explanation, and a question asking the model to improve its response.
	\item We construct dialogues of 1-3 turns, ensuring that the final ground-truth answer is always the highest-scoring candidate.
	\item For multi-turn cases, the first-turn response is always the lowest-scoring candidate, enabling a progressive refinement process.
\end{itemize}
This answer-critic iterative refinement encourages the model to learn from mistakes, gradually correcting its responses over multiple turns.
For the rule-based rewards, there is no reasoning provided, and the scoring criteria differ from those of VLM-based rewards, which leads to a different interpretation of the relative score differences. Therefore, each round’s prompt is selected from a predefined prompt pool that expresses the meaning of \textit{“could you generate a better answer.”}

\begin{table*}[t]
\centering
\caption{Comparison of RePer’s Performance with Baselines and State-of-the-Art Models on the GAPE Benchmark.}
\label{tab-exp:gape}
\vspace{3mm}
\resizebox{0.7\textwidth}{!}{ 
\begin{tabular}{l|cccccc}
\toprule
\textbf{Model} & \multicolumn{6}{c}{\textbf{GAPE}}\\
               & Authenticity ↑ & Correctness ↑& Detail ↑ & Coherence ↑ & Completeness ↑ & Total ↑\\
\midrule

LLaVA-SFT+ 7B              & 27.62 & 12.47 & 12.27 & 9.61 & 8.11 & 70.09\\
LLaVA-RLHF 7B              & 27.93 & 12.64 & 12.44 & 9.55 & 8.11 & 70.68\\
VOLCANO 7B                 & 31.63 & 14.52 & 13.89 & 9.86 & 8.90 & 78.78\\
LLaVA-SFT+ 13B             & 30.00 & 13.44 & 13.09 & 9.76 & 8.58 & 74.88\\
LLaVA-RLHF 13B             & 30.06 & 13.59 & 13.39 & 9.71 & 8.61 & 75.36\\
VOLCANO 13B                &31.34	&14.32	&13.76	&\textbf{9.85}	&8.9	&78.17\\
\midrule
LLaVA-1.5 7B               & 30.19 & 13.58 & 13.15 & 9.78 & 8.46 & 75.16\\
 \rowcolor{mydred} \textbf{ +RePer} & 33.16 & 14.95 & 13.95 & 9.87 & 8.96 & 80.88\\
\midrule
LLaVA-1.5 13B              & 31.27 & 14.12 & 13.48 & 9.81 & 8.69 & 77.37\\

\rowcolor{mydred} \textbf{ +RePer}        & \textbf{34.11} & \textbf{15.33} & \textbf{14.26} & 9.70 & \textbf{9.15} & \textbf{82.54} \\
\bottomrule
\end{tabular}
}
\end{table*}

\section{Details of GAPE}
\label{sup:gape}
As introduced in~\cref{exp:gape}, GAPE employs evaluation prompts aligned with the “Rating Criteria” outlined in \cref{fig:data-reward-pipe}. The evaluation score ranges from 0 to 100, with the following weight distribution: Authenticity (40 points), Correctness (20 points), Detail (20 points), Coherence (10 points), and Completeness (10 points).

GAPE serves as a complement to traditional image captioning benchmarks. Existing benchmarks, such as COCO Caption~\cite{chen2015microsoft} and NoCaps~\cite{agrawal2019nocaps}, rely on language metrics like BLEU, ROUGE, and CIDEr, which are constrained by closed-set vocabularies and require human-annotated ground truth captions, making them less scalable for evaluating vision-language models (VLMs) that generate diverse descriptions. In contrast, GAPE provides a more flexible and robust evaluation pipeline, operating without human-annotated ground truth and leveraging LLM-based assessment to better align with human judgment, while accommodating the open-ended nature of caption generation.

\cref{tab-exp:gape} presents a detailed comparison of model performance on GAPE across all evaluation aspects.

\section{Case Study}
\label{sup:case}
\begin{figure*}[t]
\centering
\includegraphics[width=1.0\linewidth]{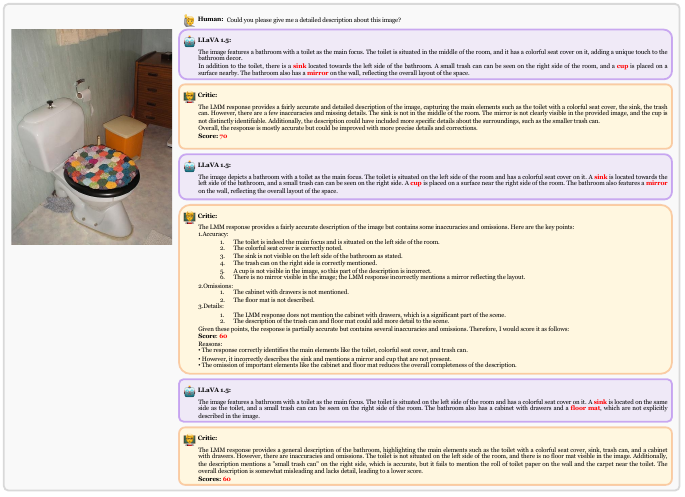}
\caption{\textbf{Failure case of LLaVA-1.5 in answer refinement}. Despite receiving critic feedback on the initial response, LLaVA-1.5 fails to correct the hallucinated objects (marked in red) in the second turn and further introduces the nonexistent “floor mat” in the third turn.}
\label{fig:case-llava-multiturn}

\end{figure*}

\cref{fig:case-llava-multiturn} illustrates a failure case of the baseline model LLaVA-1.5~\cite{llava1p5} in a multi-turn conversation with a critic model (LLaVA-Critic~\cite{llavacritic}). In this setup, LLaVA-1.5 first generates an initial response and then attempts to refine its answer based on feedback from the critic. Despite receiving detailed and accurate feedback, the model fails to correct its mistakes in the second turn, continuing to include hallucinated objects (highlighted in red) from the initial response. Furthermore, in the third turn, it introduces additional hallucinated objects, demonstrating its inability to properly leverage feedback for refinement. 
This failure also results in stagnant or even lower scores assigned by the critic.
This underscores the necessity of equipping models with stronger feedback-driven self-correction mechanisms to improve response accuracy and mitigate hallucinations.

\begin{figure*}[t]
\centering
\includegraphics[width=1.0\linewidth]{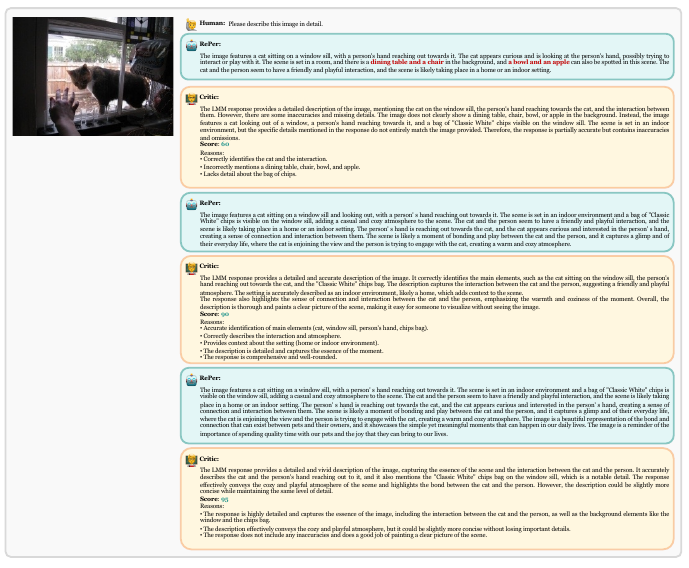}
\caption{\textbf{Successful refinement case of our RePer}. Initially, RePer generates hallucinated objects but corrects them in the second turn based on critic feedback. By the third turn, it enhances the description with more details, improving a response that already scored 90. The increasing feedback scores highlight RePer’s iterative refinement and progressive improvement.}
\label{fig:case-reper-multiturn}
\end{figure*}

\begin{figure*}[t]
\centering
\includegraphics[width=0.9\linewidth]{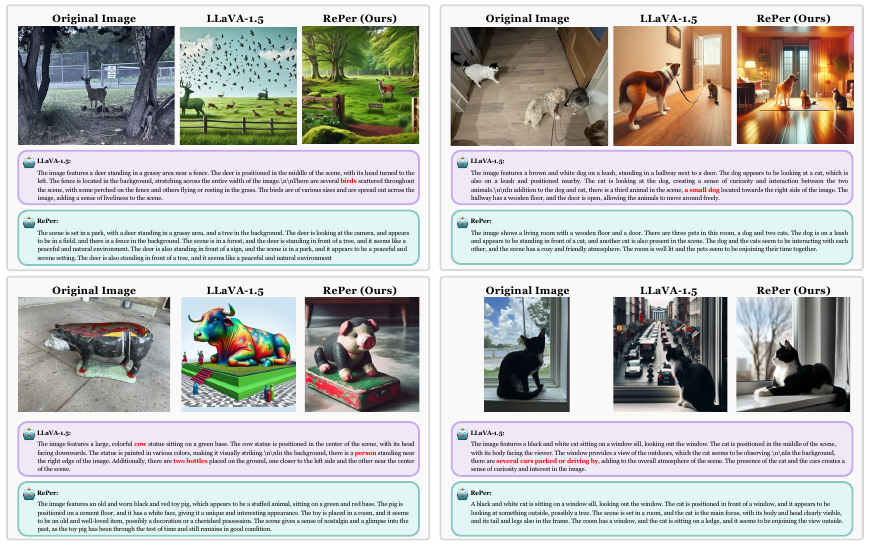}
\caption{\textbf{Comparison of initial-round outputed image captions between the baseline model and RePer.} RePer generates more accurate and detailed descriptions, while the baseline model often introduces hallucinated objects. The corresponding DALLE-3 generated images further illustrate that RePer’s captions lead to more faithful image reconstructions.}
\label{fig:case-dalle-caption}
\end{figure*}

\begin{figure*}[htbp]
\centering
\includegraphics[width=0.9\linewidth]{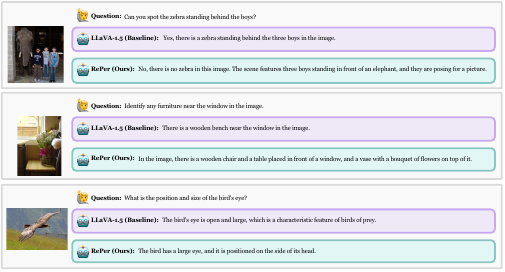}

\caption{\textbf{Question-answering cases evaluating models' image understanding.} Responses are generated in a single-turn format without iterative refinement.}
\label{fig:case-qa}
\end{figure*}

\cref{fig:case-reper-multiturn} illustrates a successful refinement case of RePer. While the initial response contains hallucinated objects, RePer corrects these errors in the second turn based on feedback from the critic model. By the third turn, it further enhances the description by adding more details to an already accurate response, which had received a score of 90 from the critic. Throughout the multi-turn conversation, the feedback score steadily increases, reflecting the model’s progressive improvement. This demonstrates RePer’s strong ability to iteratively refine its answers based on prior responses and feedback, effectively guiding itself toward a more detailed and accurate final answer.

\cref{fig:case-dalle-caption} compares the initial-round outputs of the baseline model and RePer, highlighting RePer’s strong ability to generate accurate and detailed descriptions. While the baseline model frequently introduces hallucinated objects, RePer consistently produces more faithful image descriptions. Additionally, we present images generated by DALLE-3~\cite{dalle3} as part of the evaluation process in \cref{exp:txt2img}. The higher-quality captions from RePer lead to synthesized images that more closely resemble the original inputs, further demonstrating its effectiveness as a good captioner.

\cref{fig:case-qa} presents question-answering cases that assess the model’s image understanding capabilities. The responses are generated in a single-turn answer format without iterative refinement.





\end{document}